# Deneb-Vritra: Descriptions, Edges and Names Examined by Blocking Via Restoring Token Representations and Attention

*Where a One-Pass Decision Encoder Reads Option Names and Descriptions. Attention knockouts, head ablations, residual patching, probes and layer grafting in Laya, with released, fine-tuned and name-randomised checkpoints*

Mohammed Nayeem Khan
mohammed.nayeem.khan@intel.com
mnknayeem@gmail.com

Akhil P
akhil.p@intel.com

Shahina Shaik
shahina.shaik@intel.com

Intel Corporation

**Abstract**

Laya is a 421M-parameter encoder that answers a question about a text in one pass: each answer option, written as a name plus a description, gets a marker token, and one score is read from the network at each marker. Earlier papers showed that such models follow an option's name when it contradicts the description, but not where in the network this happens. We locate it. On 951 synthetic questions we move each name onto another option's description and count how often the model picks the option with the right name (the name shortcut): 53% for the released checkpoint, 59% for our fine-tune v1, 11% for a fine-tune with name randomisation. On a 320-item sample, blocking attention from each marker to its own name tokens in layers 14 to 17 of 28 cuts v1's shortcut from 67% to 16% (19% on two schemas held out from component selection) while accuracy with the original names stays at 90%; blocking marker edges to the question, other options or text has no effect. No single head matters (largest effect 9 points), and restoring clean marker states undoes the swap from layer 16. Attention mass tracks the window by layer but not by head. Repeated on 5 checkpoints (released, fine-tuned, name-randomised, and a pair trained without two schemas), the knockout finds similar layers; in the name-randomised ones, blocking the description edges instead brings the shortcut back to 56-58%, so the read persists and is outvoted. Marker states decode the name from layer 0; a one-direction probe erasure changed nothing but was weaker than a random edit. Copying layers 0-13 or 14-27 of a name-randomised sibling into a fine-tune recovers 66% and 65% (v1); 21% and 45% (pair A) of its shortcut reduction. 9 of the 12 predictions we could score, written before running, failed. Head sweeps cover 5 of 5 checkpoints; data are synthetic and LLM-labelled; one architecture.

**Key findings (every number recomputed at build time)**

1. The name shortcut of v1 is read by marker-to-name attention edges in layers 14 to 19: blocking them in layers 14 to 17 lowers the swap shortcut on the 320-item sample from 66.9% to 16.2% [12.6, 20.7], with accuracy for original names at 90.0% [86.2, 92.8] against 90.3% (Section 6).
2. On the schemas held out from component selection the shortcut falls from 75.0% to 19.4% [14.0, 26.2]; blocking the marker-to-description edges instead raises it (Section 6).
3. The effect is spread over many heads: the largest single-head effect is 8.8 points and the ten best heads hold 29% of the summed effect (Section 7).
4. Residual patching agrees: marker states restore the correct-description margin from layer 16 (selection) and 16 (held-out); description, state and question tokens restore nothing (Section 8).
5. Head-averaged attention mass tracks the causal window across layers (Spearman 0.73) and not across heads (0.01) (Section 5).
6. Fine-tuning does not move the behaviour much (released 52.7%, v1 58.8% on all 951 questions), name-randomised fine-tuning does (10.9%) (Section 4).
7. 9 of the 12 pre-registered predictions that could be scored failed, among them the location of the name read and the sparsity of the heads involved (Section 13).
8. Not every checkpoint received every experiment (Section 14).

## 1 Introduction

A typed-decision model answers a bounded question about a text in one forward pass and generates no text. The caller sends a text (the state), a question, and K options. Each option is a short name plus a one-sentence description. Laya, a 421-million-parameter model built on ModernBERT-large [1], writes each option into its input as “name: description”, puts one [MASK] marker token in front of it, and reads one score from the encoder at each marker. The K scores become K probabilities.

The description is where a caller states the rule, so the interface implies that the probabilities follow the descriptions. Two recent papers report that Laya and related models follow the name when the two disagree [2, 3]. We reproduce the effect with a conflict test on 951 synthetic questions: every option name is moved onto another option’s description, and the description of the correct option stays fixed. The released Laya checkpoint then picks the option that carries the correct *name* on 52.7% [49.5, 55.8] of the questions and the option with the correct *description* on 29.8% [26.9, 32.7]. Our fine-tune of it (called v1 below) behaves the same way (58.8% [55.6, 61.9] and 33.1% [30.2, 36.2]). A second fine-tune that saw name randomisation (v1-rename) follows the name on 10.9% [9.1, 13.1].

These studies are behavioural. Azizi et al. [3] change how the option string is rendered and show that the effect vanishes when the name is left out of the input, so they locate the cause in the input format. They do not open the network. Three questions remain. Which layers and which tokens carry the name into the decision? Is the wiring in the released checkpoint the same as in a fine-tuned one and in a name-randomised one? Does name randomisation remove the wiring, or does it leave the wiring in place and add a stronger competitor?

We answer with causal tools on one harness: attention knockouts (blocking chosen token-to-token edges in a window of layers), mean-ablation of single heads and of whole layers, residual-stream patching between the original and the swapped input, linear probes on marker states, and layer grafting between checkpoints. The harness runs the encoder in fp32 with the attention function replaced; its logits match the stock forward pass to 6e-5 and its answers match the Laya agent on every item we compared (Section 3.5). Before each experiment we wrote down a numerical prediction from the layout of the input and from what is known about BERT-style encoders (Section 13). Components are selected on two schemas and tested on two other schemas, and the conflict operator used for selection (a cyclic swap) differs from two of the operators used for testing.

Our main results, in the order of the paper:

- Behaviour (Figure 2). The name wins in most conflicts for the released and fine-tuned models, and deleting every description costs v1 only 2.6 accuracy points.
- Attention knockouts (Figure 7). The name shortcut of v1 runs through edges from each option's marker to that option's own name tokens in a window of a few layers in the middle of the encoder. Blocking them there removes most of the shortcut and leaves accuracy with the original names unchanged. Blocking the marker's edges to the question, the other options or the state does nothing.
- Heads and patching (Figure 14, Figure 16). No single head is responsible, the effect is spread over dozens of heads in the same layers, and restoring the marker states from the original input undoes the corruption exactly when the window closes.
- Checkpoints (Section 10). Blocking the marker-to-name edges in the best four-layer window changes the swap shortcut rate as follows: released Laya 54.1% to 28.1% (layers 14 to 17); v1 (fine-tuned) 66.9% to 15.0% (layers 16 to 19); v1-rename (name-randomised) 11.2% to 4.7% (layers 14 to 17); pair A: fine-tuned 64.4% to 17.5% (layers 14 to 17); pair A: name-randomised 15.9% to 8.4% (layers 14 to 17). Section 10 compares the checkpoints.

Of the 12 pre-registered predictions that could be scored, 9 failed (Section 13). The largest miss is the location: we expected name reading in the first third of the network and found it in layers 14 to 17 of 28.

## 2 Related work and what is new

### 2.1 Typed-decision models and the name-definition conflict

Sun et al. [2] show that decisions of Jev and two open models flip when the mapping between option names and definitions is changed, with yes/no names producing flip rates up to 70.4 points above a 0/1 control. Azizi et al. [3] study four open typed-decision models, among them three Laya checkpoints. They find that deleting all definitions leaves accuracy unchanged, that renaming options to A and B raises accuracy on their routing suite by 15.1 points, and that the effect disappears in all three Laya checkpoints when the label is removed from the rendered option string. They attribute the failure to the input format and argue against an earlier attribution to the decision head. Other typed-decision papers concern wording augmentation [4], joint versus per-option reading of options [5], order-invariant heads [6] and calibration and framing sensitivity [7]. No paper we found opens the model: we ran title and abstract searches on Hugging Face Papers and arXiv (Appendix C) and found no layer, head or activation analysis of a typed-decision encoder.

### 2.2 Tools for locating computation

Causal tracing and activation patching ask which hidden states restore a prediction [8, 9], and Zhang and Nanda [17] show that the choice of metric and corruption can change the conclusion. Attention knockout instead blocks edges between positions in a window of layers and asks where information must flow [10]; Wang et al. [11] use the same idea to show that label words collect information in shallow layers and feed the prediction in deep ones. Circuit analyses find sparse sets of heads in decoder models [14, 13], and automated methods search for them [15, 16] or trace routes by attribution [18]. Wiegreffe et al. [12] localise the answer-symbol step of multiple-choice question answering to a few middle layers and sparse heads. Their setting is the closest to ours in task shape, but it is autoregressive, it concerns the choice of a symbol, and it has no competing name channel.

For encoders, work on BERT looked at attention patterns [19], at where linguistic information sits across layers [20], at how many heads can be removed [21, 22] and at how much head attention varies [23]; Rogers et al. [24] survey it. Probing classifiers [25] need control tasks to be interpreted [26]. Whether attention weights explain predictions is contested [27, 28, 29, 30]; we report both attention statistics and interventions and compare them.

### 2.3 Fine-tuning and mechanisms

Fine-tuning has been reported to strengthen an existing circuit [31], to add a thin wrapper over capabilities that stay in place [32], and to bypass rather than delete a capability [33]. Skill-localisation by grafting a small parameter subset onto the pre-trained model [34] is the closest relative of our grafting experiment. Shortcut learning [35] and token or label bias in multiple-choice readouts [36] are the general phenomena of which the name shortcut is an instance.

### 2.4 What is new and what is not

The tools are standard. The setting is new in four ways: a bidirectional encoder whose decision is read at inserted marker tokens, two competing channels (name and description) inside each option, a released checkpoint compared with a fine-tuned and a name-randomised descendant, and selection of components on some schemas with testing on others. Within our search, the localisation of the name shortcut to specific layers, tokens and edges of a typed-decision encoder is new. We claim it only for Laya and only on synthetic items (Section 14).

## 3 Setting, data, models and interventions

### 3.1 The model and where its decision is read

Laya turns each request into one sequence: `[CLS] choice question: ... [SEP] [MASK] name: description [MASK] name: description ... [SEP] state [SEP]`. A ModernBERT-large encoder [1] with 28 layers and 16 heads of 64 dimensions processes the sequence. Every third layer (0, 3, ..., 27) uses global attention; the others use a sliding window of 128 tokens, that is 64 tokens to each side. Our inputs have a mean of 101 and a maximum of 249 tokens, so the local layers already see most of the sequence. Two pre-norm transformer layers with no positional encoding (the option head) then mix all token states, and a small MLP maps the head output at each [MASK] position to one score. A softmax over the K scores gives the probabilities. We call the [MASK] token of an option its *marker*. In the figures we shade global layers in grey.

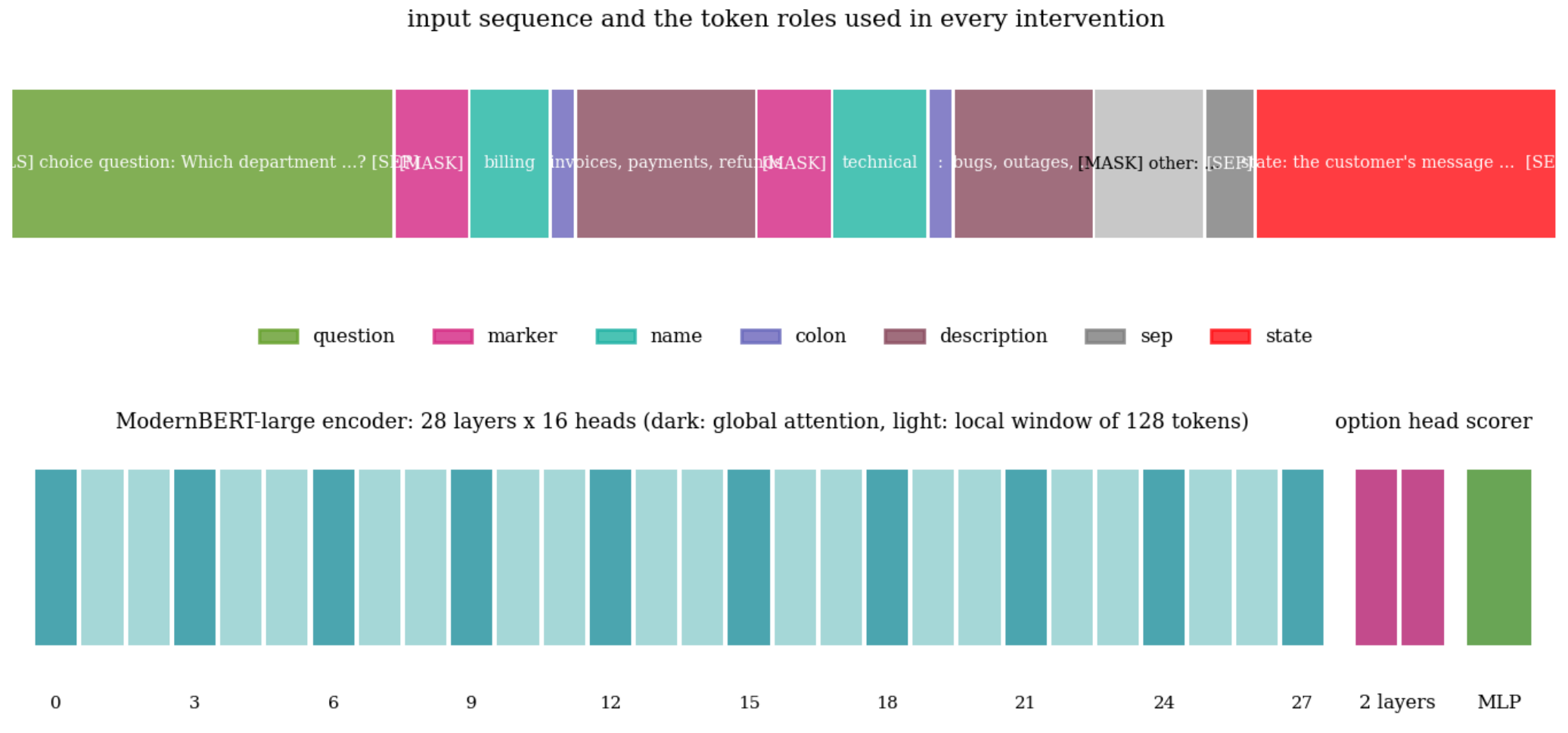

**Figure 1.** Layout of one input and of the network. Token roles (marker, name, colon, description, question, closing separator, state) define the edges we block, ablate and patch. The option head and the scorer are applied at every position; only the marker positions are scored.

### 3.2 Data

All items are synthetic. A 27-billion-parameter generator (Qwen3.8-27B) wrote texts conditioned on a target label and schema, a second blind pass of the same model labelled each text without seeing the target, and every disagreement was dropped. Test texts were generated separately and de-duplicated against the training files. An independent labeller (gemma4-31b) re-labelled the held-out items blind and agrees with the stored label on 94 to 99% of the items, depending on the set. No remote model was available for this study, so we built no new labelled data: every conflict is made programmatically from existing test rows by changing only the option names.

The conflict pool is every choice question with at least two options in the seven held-out sets: 951 questions on four schemas. The department schema (billing, technical, other) has 439 questions, the twelve-way topic schema 270, the five-way hardware failure schema 115, and the sentiment schema (positive, negative, neutral) 127. We call department and topic the *selection schemas* and failure and sentiment the *held-out schemas*. Departments, topics and sentiments have short, common-word names; the failure names include `data_corruption`. In the topic schema each description repeats its name, which makes a swap a clean test of name against description.

### 3.3 Models

The released checkpoint ("released Laya") is `convaiinnovations/laya`. v1 is our fine-tune of it with the Laya trainer (RLCD loss, 4 epochs, effective batch 64, encoder learning rate 2.5e-5, head learning rate 1e-4, option shuffling on) on 2,818 synthetic rows. v1-rename uses the same recipe on the same rows plus renamed copies of 30% of the choice rows: the names become letters, numbers or random three-letter strings, or are shifted cyclically onto other descriptions, so that the description decides. Pair A repeats the plain and the name-randomised recipe on a training file from which the failure and sentiment schemas were removed, so the held-out schemas are new to those two models. Seeds of v1 and v1-rename are used where noted.

**Table 1.** What fine-tuning and testing share. All test states are disjoint from all training files. Held-out schemas apply to pair A only; for the other models the 'held-out' split is a selection/test split of components, not of training data.

| Model | Fine-tuning rows | Schemas seen | Names and descriptions | Shift used in training | Conflicts used at test |
|---|---|---|---|---|---|
| Released Laya | not trained by us | unknown to us | n/a | unknown | swap, pair, derangement |
| v1 | 2,818 | department, failure, topic, sentiment | identical to test | none | swap, pair, derangement |
| v1-rename | 3,579 | same four | identical, plus letters, numbers, random strings | cyclic shift of names (one of four schemes) | swap (in-family), pair, derangement (not cyclic) |
| Pair A, fine-tuned | 2,165 | department, topic only | identical to test for those two | none | swap, pair, derangement |
| Pair A, name-randomised | 2,730 | department, topic only | plus renamed copies | cyclic shift of names | swap (in-family), pair, derangement |

### 3.4 Conflicts and measures

For each question we keep the state, the instruction and all descriptions, and change the names. *Swap* moves each name to the next option's description (cyclic, offset 1; this is one of the shifts used when training the name-randomised models). *Pair* exchanges the first two names, and only questions whose correct option is one of the two are scored. *Derangement* applies a random permutation with no fixed point that is not a cyclic shift (K at least 4: the failure and topic schemas). Two controls leave the conflict out: *nodesc* deletes every description, *letters* replaces names by A, B, C.

We report two rates per condition. *Description accuracy* is the share of items for which the highest-scored option is the one whose description is correct. The *name-shortcut rate* is the share for which the highest-scored option is the one that now carries the correct name, counted only when that option differs from the right one. For continuous effects we use the *name gap*, the score of the name-carrying option minus the score of the description-correct option (positive means the name leads; scores are raw logits before any calibration temperature), and the *margin* of the correct option over its best competitor. Intervals are Wilson 95% intervals for rates and bootstrap 95% intervals over items for differences and means.

### 3.5 Harness and interventions

The harness rewrites the encoder's attention function (eager attention with hooks) and runs everything in fp32. On 48 items its logits differ from the stock forward pass by at most 6e-5 and the argmax agrees on all of them, and its answers equal the Laya agent's (which runs in bf16) on the same 48 items. Patching a layer with its own activations changes nothing (difference 0.0), and the attention rows of every head sum to one.

*Attention knockout* sets the attention logits of chosen (query, key) pairs to minus infinity in every head of a window of layers. A specification names the query role, the key role and a relation (the same option, other options, or anything). *Ablation* replaces the output of one head (or all heads of one layer) by its mean over all token positions of 80 original-name items (zero ablation is a robustness check). *Patching* copies residual-stream states at chosen positions from the original-name run into the swapped-name run at the output of one layer; restoration is (patched margin minus swapped margin) divided by (original margin minus swapped margin), as in [17]. *Probes* are ridge classifiers on standardised marker states (control tasks follow [26]). *Grafting* copies the parameters of a layer range from one checkpoint into another [34].

Shared sample. All interventions run on a fixed sample of 80 questions per schema (320 swap items; 136 pair items; 160 derangement items). Heads and layers are ranked on 80 swap items from the selection schemas (a separate random draw from the same pool) and evaluated on the full sample, with the held-out schemas reported separately. Where an experiment has more than one conflict operator we say which one was used for selection.

## 4 Behaviour: how often the name wins

Table 3 and Figure 2 give the baseline. With the original names the released checkpoint is right on 69.9% [66.9, 72.8] of the 951 questions, and v1 on 86.4% [84.1, 88.5]. Fine-tuning on our schemas thus lifts accuracy by about 17 points, but it does not change who wins a conflict. Under the cyclic swap the released checkpoint follows the name on 52.7% [49.5, 55.8] of the items and v1 on 58.8% [55.6, 61.9]; the paired difference is +6.1 points [+2.4, +9.8], so fine-tuning slightly strengthened the shortcut. Across the three further seeds of v1 the cyclic-swap shortcut rate lies between 47.8% and 58.8%.

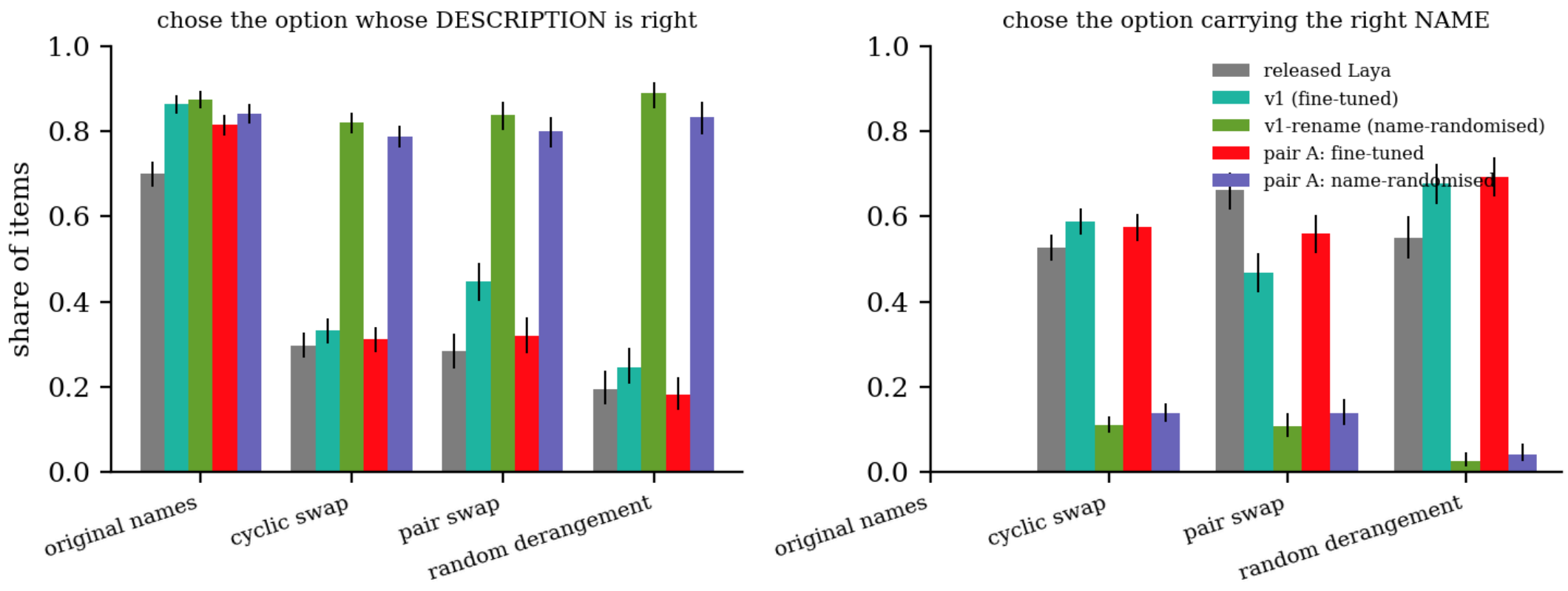

**Figure 2.** Behaviour of the five main checkpoints. Left: the option with the correct description is chosen. Right: the option that carries the correct name is chosen (cyclic swap, pair swap, random derangement). Bars show Wilson 95% intervals.

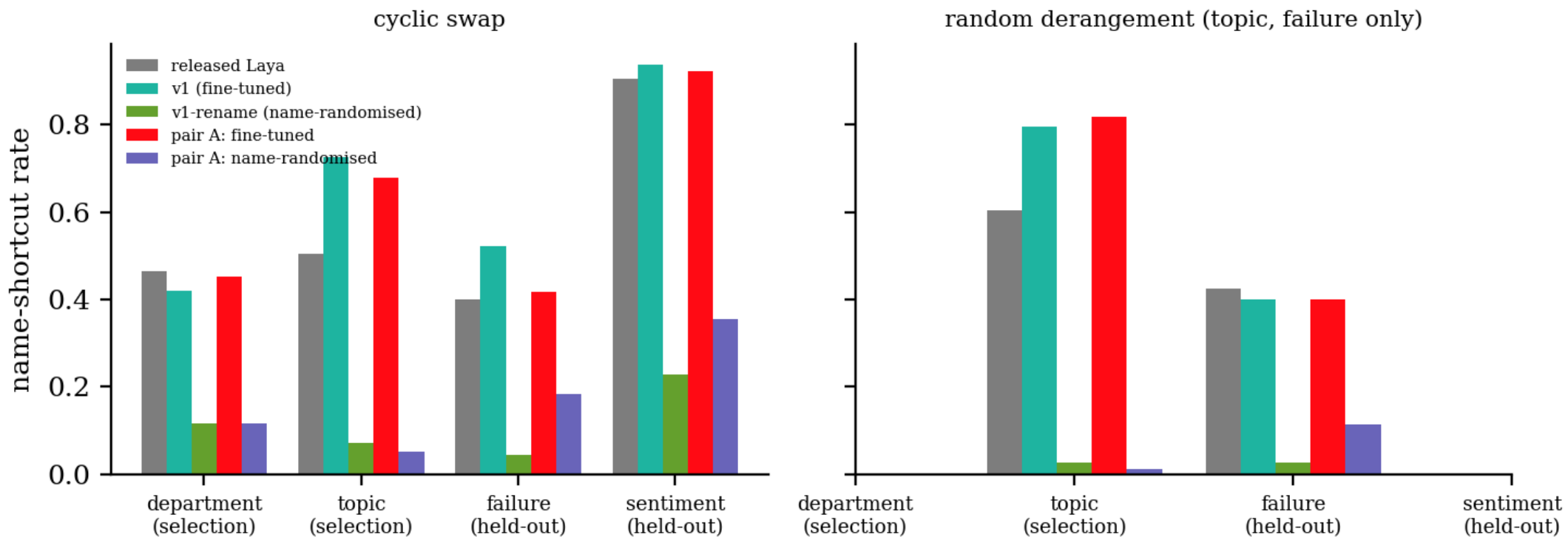

**Figure 3.** Name-shortcut rate by schema for the five main checkpoints. Department and topic are the selection schemas; failure and sentiment are the held-out schemas (new to pair A).

Figure 3 splits the cyclic-swap shortcut by schema: v1 follows the name on 41.9% (department), 72.6% (topic), 52.2% (failure), 93.7% (sentiment) of the items.

**Table 3.** Behaviour on the conflict pool (percent of items; n = 951 for orig, swap, nodesc and letters, 466 for pair, 385 for derangement). Intervals are in Appendix Table 2.

| Model | orig names | swap: description right | swap: name followed | pair: name followed | derangement: name followed | no descriptions | letters as names |
|---|---|---|---|---|---|---|---|
| released Laya | 69.9 | 29.8 | 52.7 | 66.1 | 55.1 | 69.6 | 68.1 |
| released typed-decisions | 71.2 | 36.8 | 47.5 | 49.8 | 50.6 | 70.8 | 69.8 |
| v1 (fine-tuned) | 86.4 | 33.1 | 58.8 | 46.8 | 67.8 | 83.8 | 85.3 |
| v1 seed 1 | 86.1 | 37.2 | 54.6 | 49.8 | 63.9 | 83.5 | 85.9 |
| v1 seed 2 | 85.8 | 44.5 | 47.8 | 36.7 | 63.1 | 82.2 | 85.5 |
| v1 seed 3 | 86.6 | 36.4 | 55.8 | 63.9 | 65.7 | 84.1 | 85.9 |
| v1-rename (name-randomised) | 87.5 | 82.0 | 10.9 | 10.7 | 2.6 | 83.0 | 86.1 |
| v1-rename seed 1 | 88.0 | 86.1 | 7.5 | 10.3 | 2.9 | 80.7 | 85.9 |
| v1-rename seed 2 | 87.4 | 81.8 | 10.1 | 8.2 | 2.6 | 81.8 | 85.4 |
| shift-only rename | 88.0 | 87.2 | 6.8 | 7.3 | 1.0 | 80.7 | 87.5 |
| neutral-only rename | 85.7 | 41.0 | 51.6 | 45.9 | 61.8 | 83.9 | 86.8 |
| pair A: fine-tuned | 81.5 | 31.1 | 57.4 | 56.0 | 69.4 | 80.4 | 81.4 |
| pair A: name-randomised | 84.1 | 78.8 | 13.8 | 13.7 | 4.2 | 80.7 | 84.4 |

Name randomisation removes most of the shortcut. v1-rename follows the name on 10.9% [9.1, 13.1] under the cyclic swap (difference to v1 -47.8 points [-51.3, -44.5]) and on 2.6% [1.4, 4.7] under random derangements, an operator that is not a cyclic shift and so was not part of its training distribution. Names replaced by neutral strings alone do not help (51.6% followed), while names shifted onto other descriptions do (6.8%); the shifted-name copies are the ingredient that teaches the model that the description decides. Seeds of v1-rename lie between 7.5% and 10.9%.

Pair A shows what happens off the training schemas. Both models were trained without the failure and sentiment schemas. On the two schemas they did train on, the name-randomised model follows the name on 9.2% [7.3, 11.5] of the swap items, against 53.7% [50.1, 57.4] for the plain fine-tune. On the held-out schemas the figures are 27.3% [22.0, 33.2] and 68.2% [62.1, 73.7]. Name randomisation therefore transfers to new schemas, but only partly: the residual shortcut is about 3.0 times larger on schemas it never saw.

The two controls tell us how much each channel carries. Deleting every description lowers v1's accuracy by 2.6 points [0.8, 4.5] and the released checkpoint's by 0.3 points [-2.2, 2.8]. On the three schemas in which descriptions add information beyond the name (department, failure, sentiment; in the topic schema the description repeats the name) the corresponding losses are 2.9 points [0.7, 5.3] for v1 and 2.8 points [-0.3, 5.7] for the released checkpoint. Replacing names by letters lowers v1's accuracy by 1.2 points [-0.5, 2.8]. In these schemas the names carry nearly all of the evidence and the descriptions add a few points, which is the behavioural counterpart of the observation by Azizi et al. [3] that deleting definitions leaves accuracy almost unchanged. It also means that a model can win the original-name test with either channel, which is why the conflict test is needed.

In-family and held-out. For v1, v1-rename and the released checkpoint the four schemas were seen (or, for the released checkpoint, are unknown), so 'held-out' in the later sections means components were chosen on the selection schemas and tested on the other two. For pair A the held-out schemas are new to the fine-tuned models as well.

## 5 Where the markers look: attention statistics

We start with what the attention weights say, because attention patterns are the cheapest evidence and a reader may take them as the answer. For each marker query we sum the attention weight that falls on each token role, average over heads and over the items of the shared sample, and compare with the share of keys that role occupies. For v1 the keys are 38.5% state tokens, 14.9% question tokens and 2.5% tokens of the marker's own name and description, so a role with more mass than share is attended preferentially. The swapped and the original inputs give nearly identical profiles (Appendix Table 4).

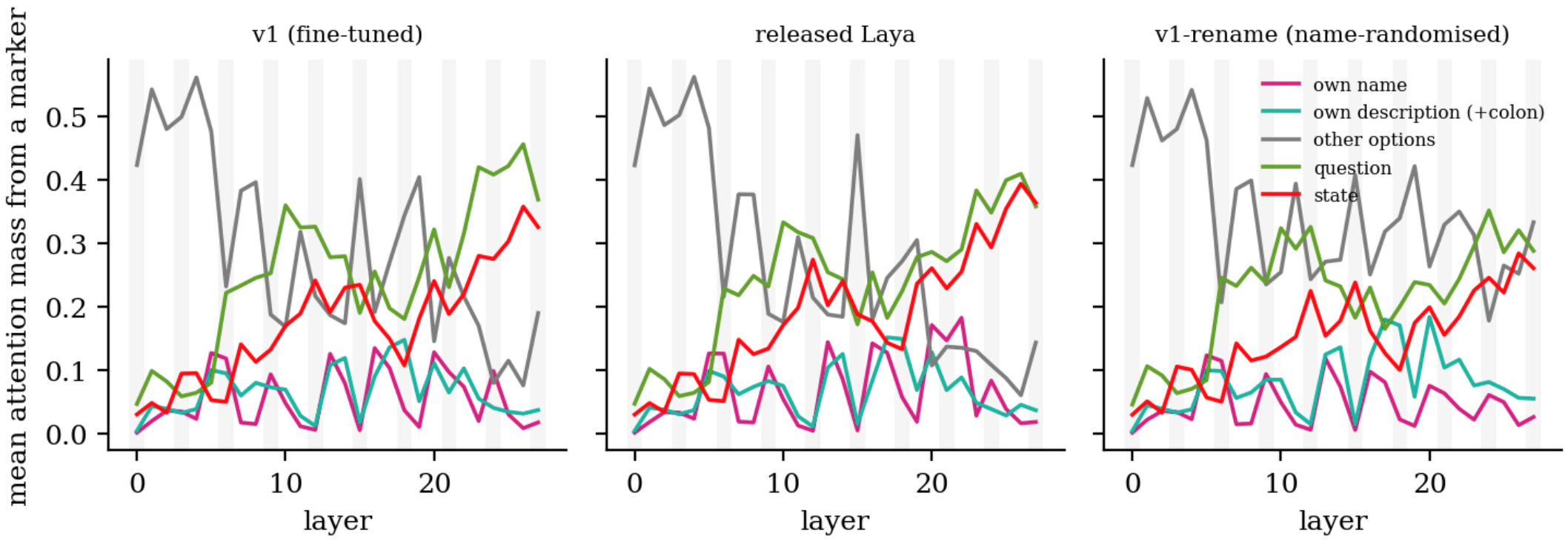

**Figure 4.** Attention mass from marker tokens to each token role, averaged over the 16 heads and over the original-name items, for v1, the released checkpoint and v1-rename. Grey bands mark global-attention layers.

Three features recur in the three checkpoints (Figure 4). In layers 0 to 3 the markers attend almost only to the other markers and to themselves: layer 0 puts 87.7% of the mass on markers in v1, 87.7% in the released checkpoint and 87.9% in v1-rename. In the middle layers the question carries 27.4% of the mass on average in v1 and the state a growing share, from 3.0% in layer 0 to 24.2% in layer 12 and 32.6% in layer 27. The marker's own name tokens receive their largest head-averaged mass in layers 5, 6, 13, 16, 17, 20 (12.3% on average, against 3.5% elsewhere); the layer profile of own-name mass in v1 correlates with that of the released checkpoint (Spearman 0.94) and of v1-rename (0.96). The description tokens receive most mass in layers 13, 14, 17, 18, 20, 22. Global and local layers do not differ much: the mean own-name mass is 5.1% in global and 5.6% in local layers, which is expected because the local window (128 tokens) covers most of a 101-token input.

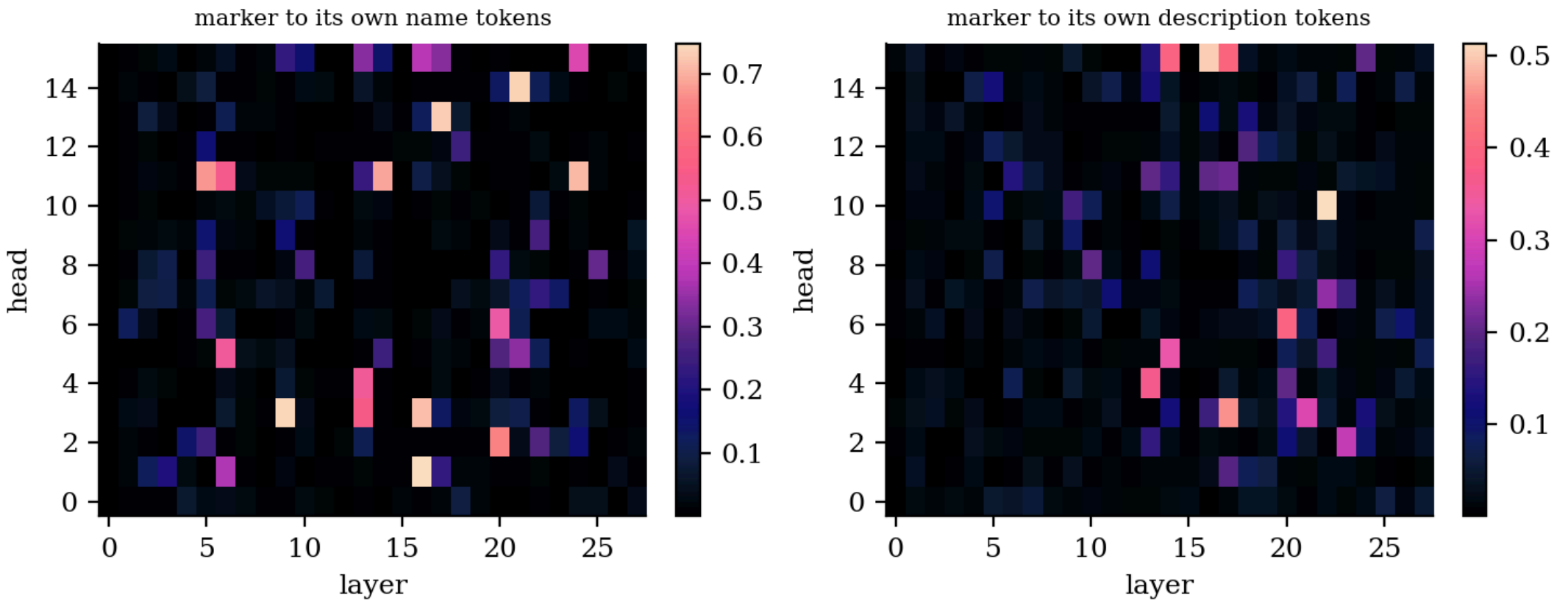

**Figure 5.** Per-head attention from markers to their own name tokens (left) and own description tokens (right) in v1. Each cell is one head of one layer.

The head-level picture is sparser (Figure 5). The strongest own-name head is layer 16, head 1, which puts 74.8% of its mass on the marker's own name; 21 heads exceed 30%. Heads with large own-description mass sit in the same layers.

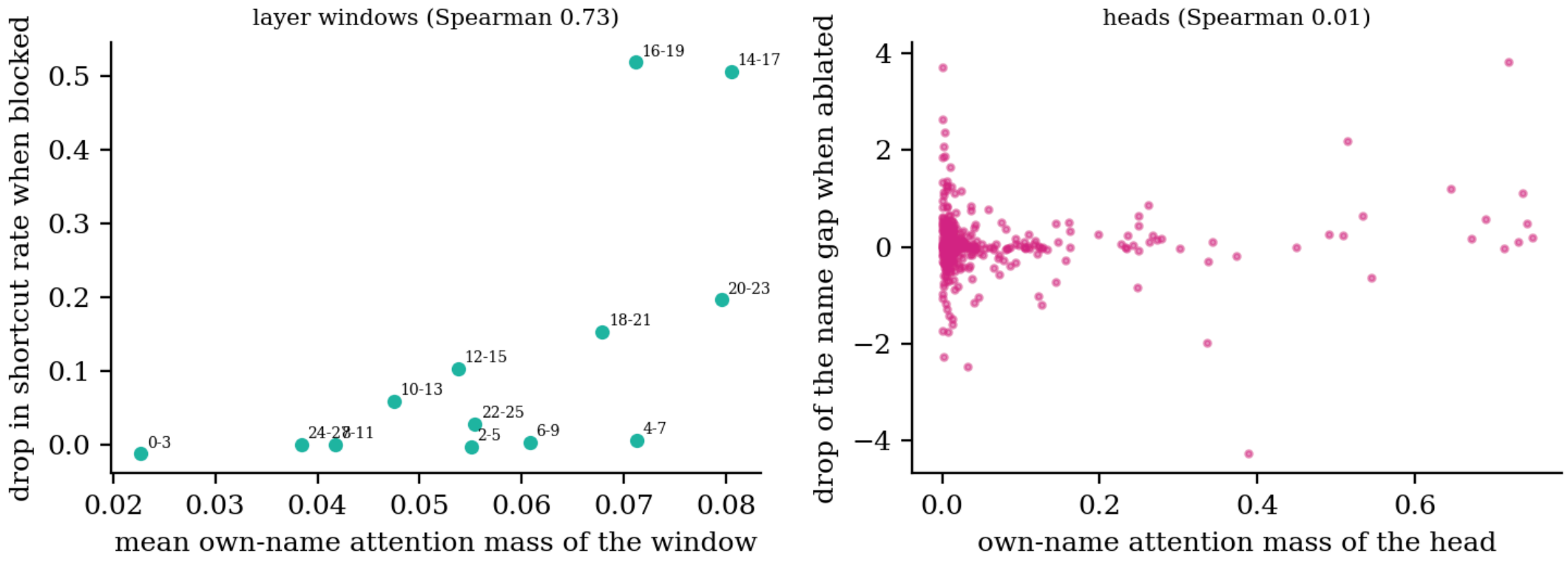

**Figure 6.** Attention against causal importance in v1. Left: own-name attention mass of a 4-layer window against the drop in the name-shortcut rate when the marker-to-name edges are blocked in that window (Spearman 0.73, 13 windows). Right: own-name attention mass of a head against the drop of the name gap when that head is mean-ablated (Spearman 0.01, 448 heads).

Attention mass is a fair guide at the level of layers and a poor one at the level of heads (Figure 6). Over the 13 four-layer windows, the average own-name mass of the window correlates with the drop in the shortcut rate that blocking the window produces with Spearman 0.73. Over the 448 heads, a head's own-name mass is uncorrelated with the effect of ablating it (Spearman 0.01, bootstrap interval -0.08 to 0.11). This agrees with the cautions in [27, 30]: weights tell where a head looks, and the interventions below tell what depends on it. The 13 windows overlap by two layers, so the layer-level correlation rests on fewer than 13 independent points and should be read as suggestive.

## 6 Attention knockouts: the reading window

### 6.1 Design

An attention knockout asks where information must travel [10]. We block edges of one kind in a window of four consecutive layers (stride two) or in all layers, in every head, and measure the swapped-name decision again. Because the knockout changes the computation rather than the input, we need controls that block edges of other kinds in the same layers. Table 5 lists the ten specifications. Layers are counted from zero and a window named by its layers covers all of them.

**Table 5.** Knockout specifications. Each one blocks, in every head of the layers of the window, the edges from the query tokens to the key tokens that satisfy the relation.

| Name | Query tokens | Key tokens | Question |
|---|---|---|---|
| marker to own name | marker | own name and colon tokens | does the marker read its option's name? |
| marker to own description | marker | own description tokens | does the marker read its option's description? |
| marker to state | marker | state tokens and closing separator | does the marker read the text? |
| marker to other options | marker | tokens of the other options | do markers compare options directly? |
| marker to question | marker | question tokens | does the marker read the instruction? |
| name to description | name and colon tokens | own description tokens | are names bound to descriptions? |
| description to name | description tokens | own name and colon tokens | are descriptions bound to names? |
| option tokens to other options | name, colon and description tokens | tokens of other options | do option tokens compare with each other? |

## 6.2 The reading window

Blocking the marker-to-own-name edges in layers 14 to 17 cuts the shortcut rate in v1 from 66.9% to 16.2% [12.6, 20.7] over the whole sample, and from 75.0% [67.8, 81.1] to 19.4% [14.0, 26.2] on the held-out schemas (paired difference +55.6 points [+48.1, +63.1]). Description accuracy under swap on the held-out schemas rises from 23.1% to 77.5% [70.4, 83.3], and accuracy with the original names stays at 90.0% [86.2, 92.8] against 90.3% (paired difference -0.3 points [-2.2, +1.6]). The windows that come within three points of the best one (the best is layers 16 to 19) cover the layers 14, 15, 16, 17, 18, 19. Windows that end before layer 14 or start after layer 19 do much less (Table 6, Figure 7). The best window is therefore in the middle of the encoder, not in its first third, where we predicted it (Section 13).

**Table 6.** Attention knockouts in v1. Name-shortcut rate under swap on the selection schemas (n = 160) and the held-out schemas (n = 160), description accuracy under swap on the held-out schemas, and accuracy with the original names on all 320 items. Wilson 95% intervals.

| Edges blocked | Shortcut, selection | Shortcut, held-out | Description right, held-out | Original names right |
|---|---:|---:|---:|---:|
| no knockout | 58.8% [51.0, 66.1] | 75.0% [67.8, 81.1] | 23.1% [17.3, 30.2] | 90.3% [86.6, 93.1] |
| marker to own name, layers 12 to 15 | 48.1% [40.5, 55.8] | 65.0% [57.3, 72.0] | 33.1% [26.3, 40.7] | 90.6% [86.9, 93.4] |
| marker to own name, layers 14 to 17 | 13.1% [8.7, 19.2] | 19.4% [14.0, 26.2] | 77.5% [70.4, 83.3] | 90.0% [86.2, 92.8] |
| marker to own name, layers 16 to 19 | 11.2% [7.2, 17.1] | 18.8% [13.5, 25.5] | 78.1% [71.1, 83.8] | 90.0% [86.2, 92.8] |
| marker to own name, layers 18 to 21 | 45.6% [38.1, 53.4] | 57.5% [49.8, 64.9] | 41.9% [34.5, 49.6] | 90.3% [86.6, 93.1] |
| marker to own name, all 28 layers | 5.6% [3.0, 10.3] | 1.9% [0.6, 5.4] | 93.1% [88.1, 96.1] | 90.6% [86.9, 93.4] |
| marker to own description, layers 14 to 17 | 81.9% [75.2, 87.1] | 90.6% [85.1, 94.2] | 3.8% [1.7, 7.9] | 88.4% [84.5, 91.5] |
| marker to own description, all 28 layers | 85.6% [79.4, 90.2] | 90.6% [85.1, 94.2] | 3.1% [1.3, 7.1] | 88.4% [84.5, 91.5] |
| marker to state, layers 14 to 17 | 58.8% [51.0, 66.1] | 76.2% [69.1, 82.2] | 21.9% [16.2, 28.9] | 90.3% [86.6, 93.1] |
| marker to state, all 28 layers | 56.9% [49.1, 64.3] | 71.2% [63.8, 77.7] | 26.9% [20.6, 34.2] | 90.3% [86.6, 93.1] |

| Edges blocked | Shortcut, selection | Shortcut, held-out | Description right, held-out | Original names right |
|---|---|---|---|---|
| marker to other options, layers 14 to 17 | 61.9% [54.2, 69.0] | 71.9% [64.5, 78.3] | 25.0% [18.9, 32.2] | 90.3% [86.6, 93.1] |
| marker to question, layers 14 to 17 | 58.1% [50.4, 65.5] | 75.0% [67.8, 81.1] | 23.1% [17.3, 30.2] | 90.9% [87.3, 93.6] |
| name to description, layers 14 to 17 | 66.9% [59.3, 73.7] | 76.9% [69.8, 82.7] | 20.6% [15.1, 27.5] | 89.7% [85.9, 92.6] |
| name to description, all 28 layers | 66.9% [59.3, 73.7] | 76.9% [69.8, 82.7] | 21.2% [15.6, 28.2] | 87.8% [83.8, 91.0] |
| description to name, all 28 layers | 66.9% [59.3, 73.7] | 76.2% [69.1, 82.2] | 21.2% [15.6, 28.2] | 89.1% [85.2, 92.0] |
| option tokens to other options, all 28 layers | 74.4% [67.1, 80.5] | 78.1% [71.1, 83.8] | 15.6% [10.8, 22.0] | 86.6% [82.4, 89.9] |

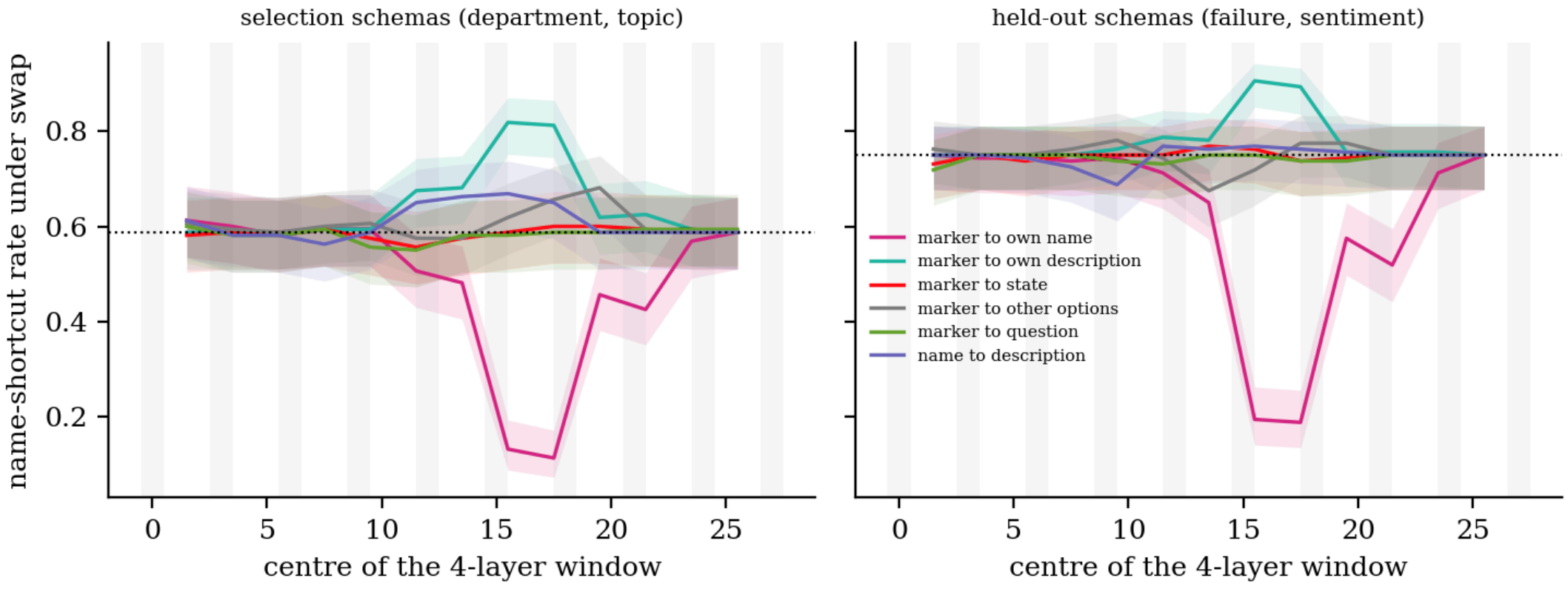

**Figure 7.** Attention knockouts in v1, by window of four layers: name-shortcut rate under swap on the selection schemas (left) and the held-out schemas (right). The dotted line is the rate without knockout; shaded bands are Wilson 95% intervals. Only the marker-to-name and marker-to-description edges matter.

### 6.3 Two channels and their controls

The description channel works against the name channel in the same layers. Blocking the marker-to-description edges in layers 14 to 17 raises the shortcut rate on the held-out schemas from 75.0% to 90.6% [85.1, 94.2], and blocking them in all layers gives 90.6% [85.1, 94.2]. A marker reads both of its option's channels in these layers and the larger signal wins. With the original names the two agree, which is why the original-name accuracy stays at 90.3% when one channel is removed.

The controls are flat. Over all windows, the largest change in the shortcut rate caused by blocking marker to state (-1.6 points at layers 10 to 13), marker to other options (+5.9 points at layers 18 to 21), marker to question (-2.8 points at layers 10 to 13), name to description (+5.0 points at layers 14 to 17), description to name (+2.5 points at layers 12 to 15), option tokens to other options (+5.6 points at layers 12 to 15) stays within 5.9 points of the 66.9% baseline, about the width of the interval on the baseline itself. Blocking marker-to-state edges in all layers leaves accuracy with the original names at 90.3%. The binding edges between names and descriptions (name to description and description to name) are not needed either, even in all layers (shortcut 71.9%, 71.6%), so the name does not travel through the description tokens on its way to the marker.

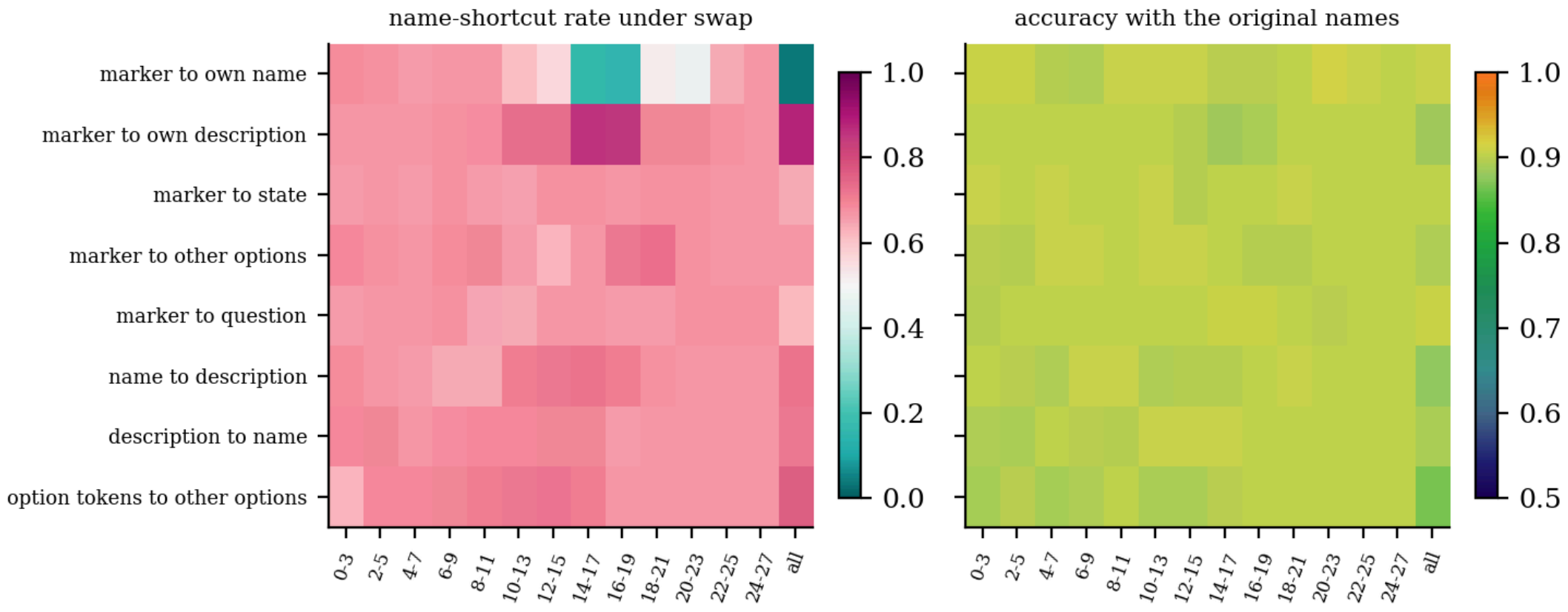

**Figure 8.** All knockout specifications of v1 by window (all 320 swap items for the left panel, all 320 original-name items for the right panel).

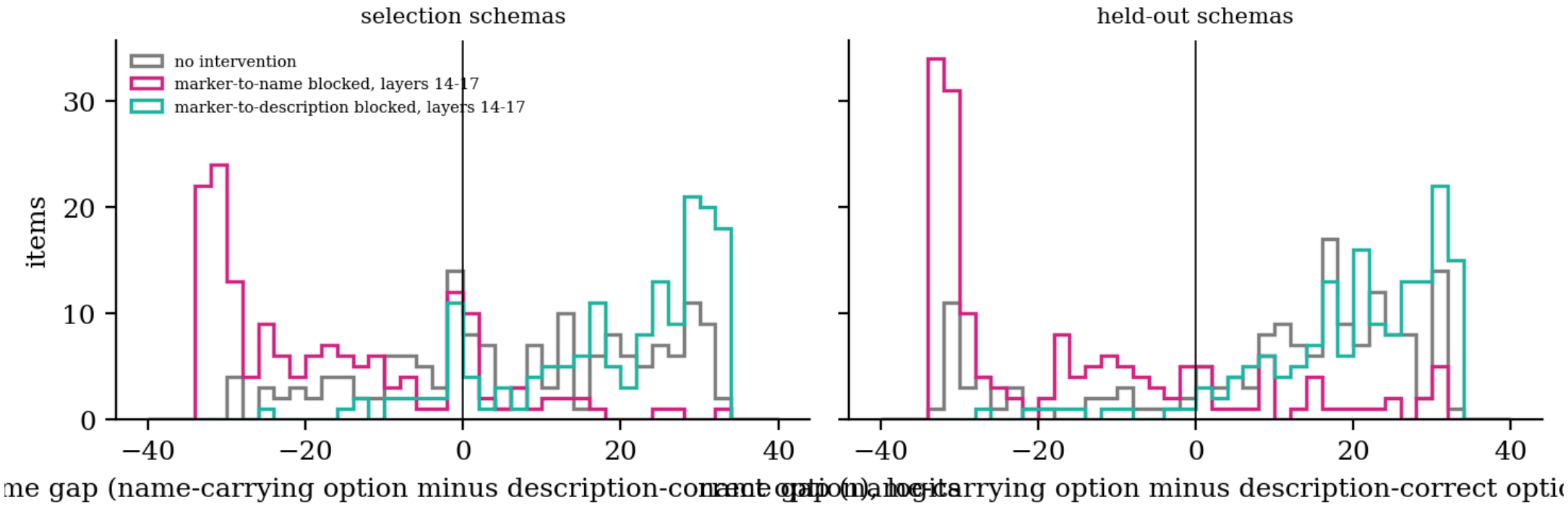

**Figure 9.** Distribution of the name gap under swap in v1, without intervention and with the marker-to-name or marker-to-description edges blocked in layers 14 to 17. Positive values: the name-carrying option leads. Values are clipped at 40 logits.

Figure 9 shows the whole distribution of the name gap. Without intervention its median is 12.2 logits; with the marker-to-name edges blocked it is -22.4, and with the marker-to-description edges blocked 23.0. The knockout moves the whole distribution rather than a tail of items.

## 6.4 Single layers

Windows of four layers hide which layers do the work. Blocking the marker-to-name edges in one layer at a time lowers the shortcut rate by at least five points in layers 13, 16, 17, 20; the largest single-layer effect is 41.2 points, in layer 16 (Figure 10). Blocking the marker-to-description edges raises the shortcut rate by at least three points in layers 13, 14, 16, 17. No single layer reproduces the effect of the four-layer window, so the name is read redundantly across several layers of the window and each layer carries part of it.

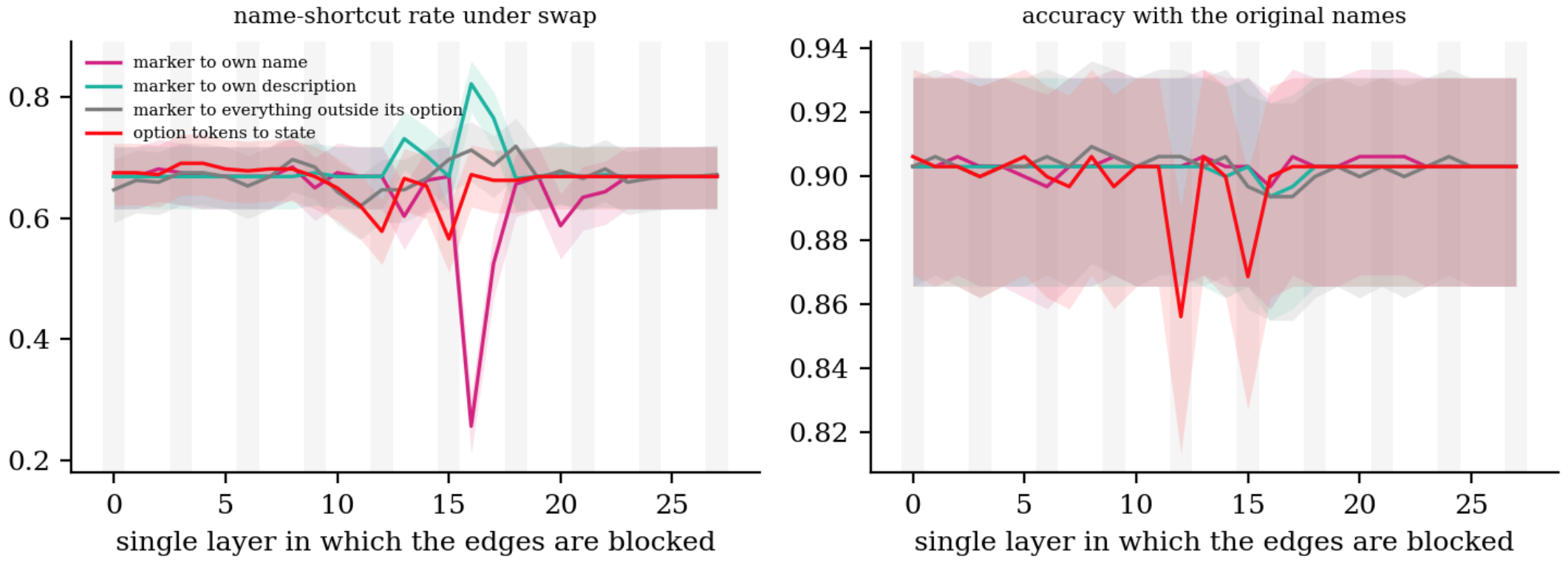

**Figure 10.** Single-layer knockouts in v1 (swap items, all 320). Left: name-shortcut rate. Right: accuracy with the original names. Four specifications; shaded bands are Wilson 95% intervals.

## 7 Heads and layers: ablation

Knockouts remove edges. Ablation removes the contribution of a component: we replace the output of one head by its mean over the token positions of 80 original-name items, and score the result on swapped inputs [21, 22]. As the effect measure we use the drop in the name gap (the score of the name-carrying option minus the score of the description-correct option, averaged over 80 swap items from the selection schemas; the mean gap is 7.5 logits before ablation). The sweep covers all 448 heads.

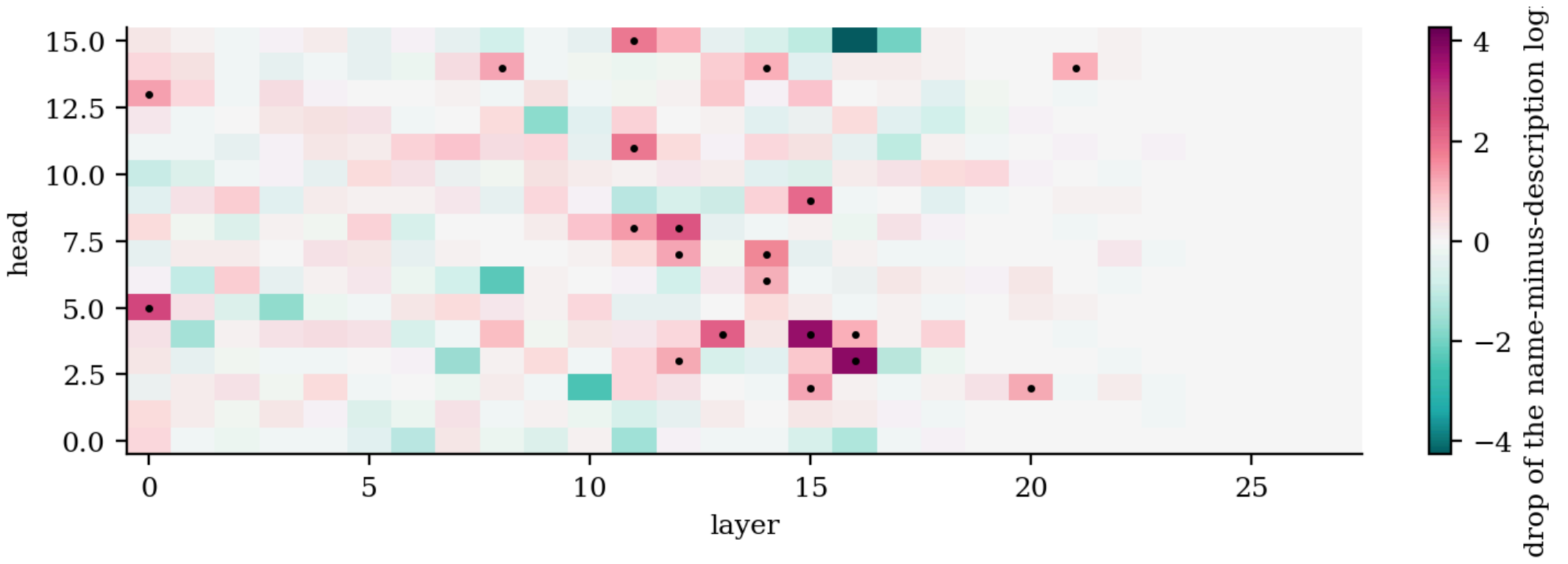

**Figure 12.** Effect of mean-ablating each head of v1 on the name gap under swap (selection schemas, 80 items). Red: the head promotes the name; blue: it works against the name. Dots mark the 20 heads with the largest effect.

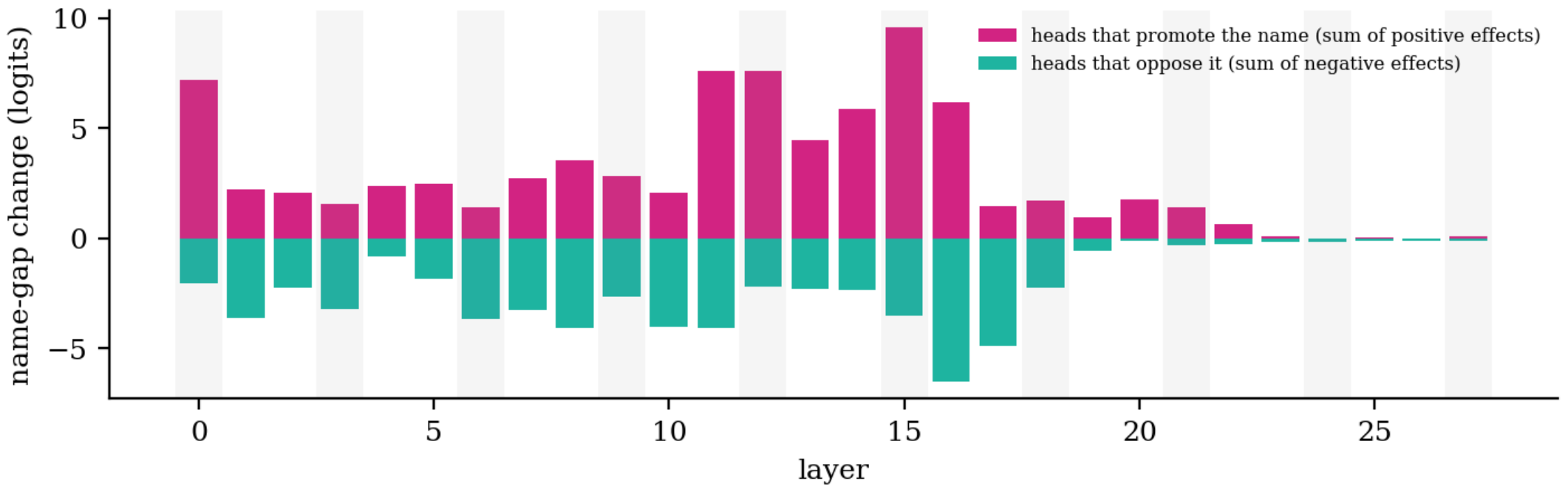

**Figure 13.** Sum over the 16 heads of each layer of the positive (red) and negative (blue) name-gap effects of single-head ablation in v1.

Summed by layer (Figure 13), the positive effects are largest in layers 15, 11, 12, 0, 16, and the negative effects (heads that work against the name) in layers 16, 17, 8.

No head is responsible alone. The largest single-head effect on the shortcut rate is 8.8 points and no head costs more than 2.5 points of accuracy with the original names (selection items, n = 80 and 40). 21 heads change the gap by more than one logit. The ten heads with the largest effect hold 29% of the summed positive effect over all heads and the top twenty 44%. Of the twenty, 15 lie in layers 11 to 17; the list is in Appendix Table 8. The distribution is wide: knocking out the marker-to-name edges in a four-layer window did far more than any set of heads we tried.

**Table 7.** Mean-ablating the k heads with the largest name effect (ranked on 80 swap items of the selection schemas). Shortcut rate on the held-out schemas under three conflict operators (n = 160 swap, about 60 pair, 80 derangement items), and accuracy with the original names on all 320 items. Random rows average five random head sets of the same size.

| Heads ablated | Swap shortcut | Pair shortcut | Derangement shortcut | Original names right |
|---|---:|---:|---:|---:|
| none | 75.0% [67.8, 81.1] | 74.3% [63.3, 82.9] | 42.5% [32.3, 53.4] | 90.3% [86.6, 93.1] |
| top 1 | 70.6% [63.2, 77.1] | 71.6% [60.5, 80.6] | 38.8% [28.8, 49.7] | 90.3% [86.6, 93.1] |
| random 1 (mean of 5) | 74.5% | 74.1% | 42.0% | 90.5% |
| top 3 | 72.5% [65.1, 78.8] | 73.0% [61.9, 81.8] | 38.8% [28.8, 49.7] | 90.3% [86.6, 93.1] |
| random 3 (mean of 5) | 72.7% | 75.4% | 41.0% | 90.1% |

| Heads ablated | Swap shortcut | Pair shortcut | Derangement shortcut | Original names right |
|---|---|---|---|---|
| top 5 | 66.2% [58.6, 73.1] | 75.7% [64.8, 84.0] | 23.8% [15.8, 34.1] | 88.8% [84.8, 91.8] |
| random 5 (mean of 5) | 72.6% | 74.6% | 43.5% | 89.7% |
| top 10 | 66.9% [59.3, 73.7] | 68.9% [57.7, 78.3] | 30.0% [21.1, 40.8] | 87.5% [83.4, 90.7] |
| random 10 (mean of 5) | 73.6% | 75.7% | 40.0% | 90.2% |
| top 20 | 65.0% [57.3, 72.0] | 60.8% [49.4, 71.1] | 23.8% [15.8, 34.1] | 87.2% [83.1, 90.4] |
| random 20 (mean of 5) | 72.5% | 73.0% | 44.7% | 89.2% |
| top 40 | 40.6% [33.3, 48.4] | 27.0% [18.2, 38.1] | 26.2% [17.9, 36.8] | 74.7% [69.6, 79.1] |
| random 40 (mean of 5) | 70.0% | 74.3% | 43.8% | 89.0% |
| top 80 | 29.4% [22.9, 36.8] | 4.1% [1.4, 11.3] | 31.2% [22.2, 42.1] | 41.9% [36.6, 47.3] |
| random 80 (mean of 5) | 66.8% | 63.2% | 45.5% | 85.2% |

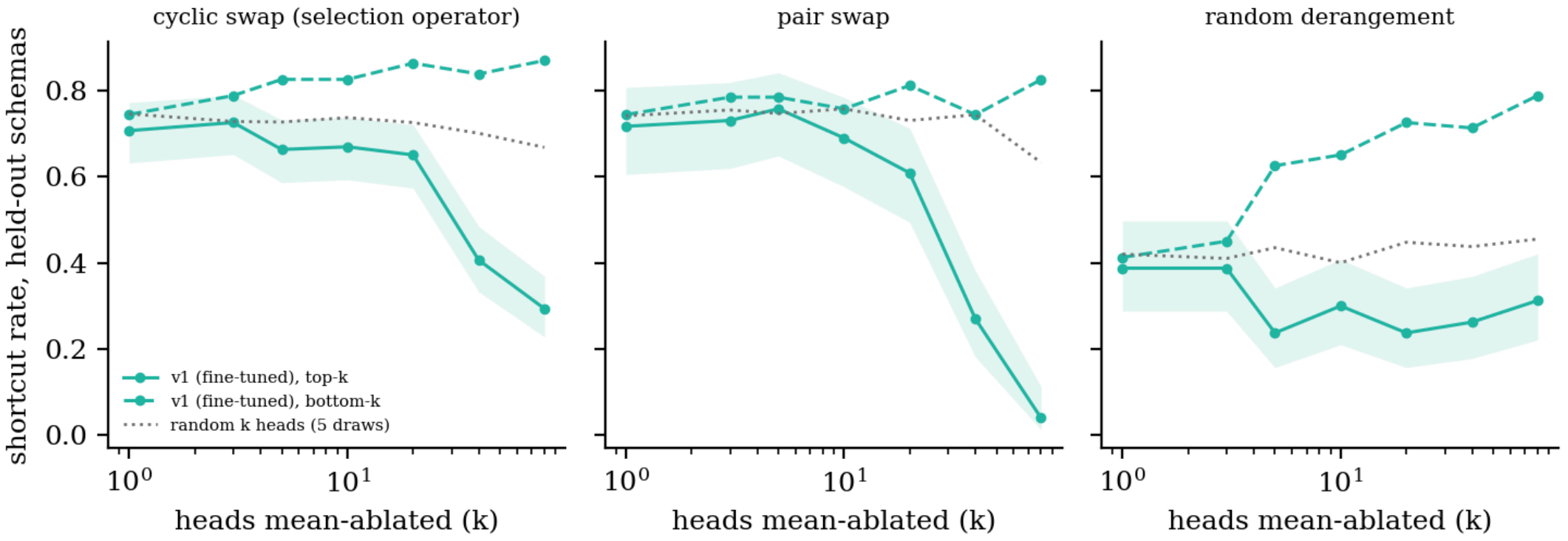

**Figure 14.** Cumulative head ablation in v1 on the held-out schemas. Solid: the k heads ranked highest on the selection schemas; dashed: the k heads ranked lowest (they work against the name); dotted: random k heads. Shaded: Wilson 95% interval of the top-k curve.

The ranking made on the selection schemas carries over to the held-out schemas and to other conflict operators, with effects that differ in size (Table 7, Figure 14). Ablating the ten top heads lowers the held-out shortcut rate under the cyclic swap, the operator used for the ranking, from 75.0% [67.8, 81.1] to 66.9% [59.3, 73.7] and the top twenty to 65.0% [57.3, 72.0]. Under the pair swap the rates go from 74.3% [63.3, 82.9] to 68.9% [57.7, 78.3] (top ten) and 60.8% [49.4, 71.1] (top twenty), and under random derangements from 42.5% [32.3, 53.4] to 30.0% [21.1, 40.8] and 23.8% [15.8, 34.1]. Ten random heads leave the swap rate at 73.6% and the ten lowest-ranked heads raise it to 82.5% [75.9, 87.6]. Accuracy with the original names is 87.5% [83.4, 90.7] after the top ten, against 90.3% [86.6, 93.1] before. With zero instead of mean ablation the top ten give a held-out swap shortcut of 65.0% [57.3, 72.0] and original-name accuracy of 86.6% [82.4, 89.9]. The cost grows with k: the top 80 heads bring original-name accuracy down to 41.9% [36.6, 47.3], so those heads do more than read names.

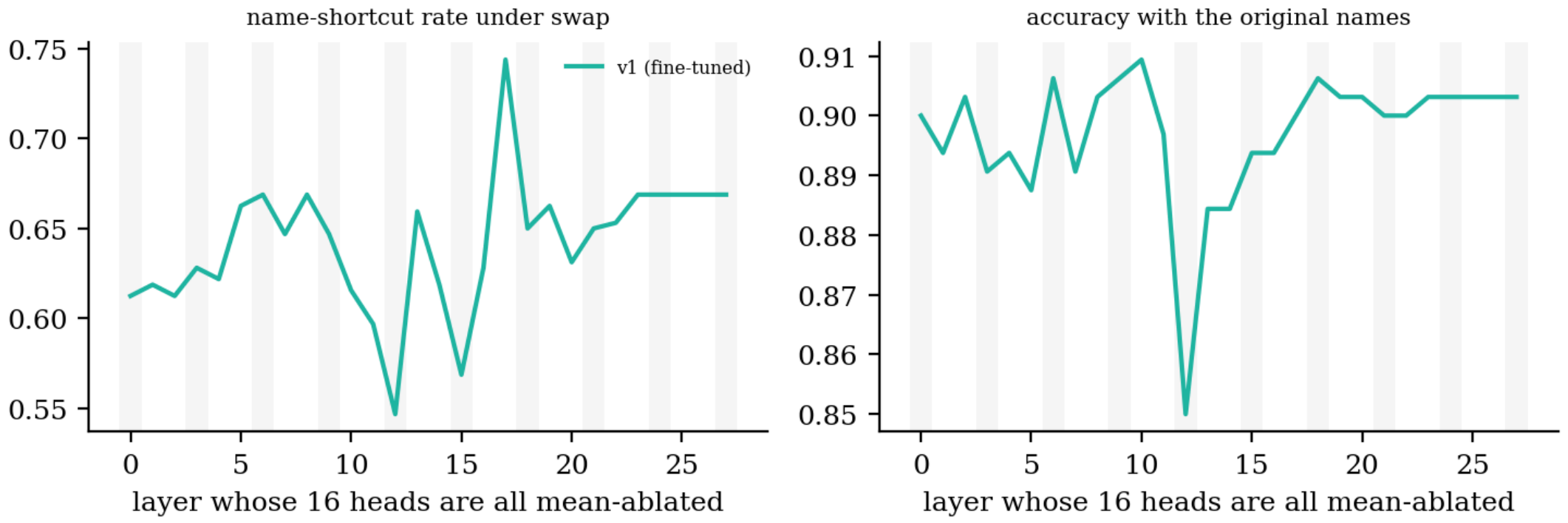

**Figure 15.** Mean-ablating all 16 heads of one layer of v1: name-shortcut rate under swap (left) and accuracy with the original names (right), on the shared sample.

Layer-level ablation (Figure 15) leads to the same region: ablating all heads of a single layer lowers the swap shortcut rate by at least five points below its 66.9% baseline in layers 0, 2, 10, 11, 12, 15, with the lowest value (54.7%) in layer 12, where accuracy with the original names is 85.0%.

## 8 Residual patching: what is sufficient and when

Knockouts and ablations tell us what the decision depends on. Patching asks what is sufficient [8, 9, 17]. We run the original-name input (clean) and the swapped-name input (corrupt) through v1, store the residual stream after each layer of the clean run, and copy selected positions of it into the corrupt run at one layer. If the copied states carry what the decision needs, the margin of the correct description recovers. Restoration is the recovered share of the gap between the two runs. The clean run is right on 90.3% of the 320 items and the corrupt run on 26.6%. Description tokens are identical in the two runs, so their positions align exactly; name spans align only for options whose name has the same number of tokens, which is 86% of the option spans.

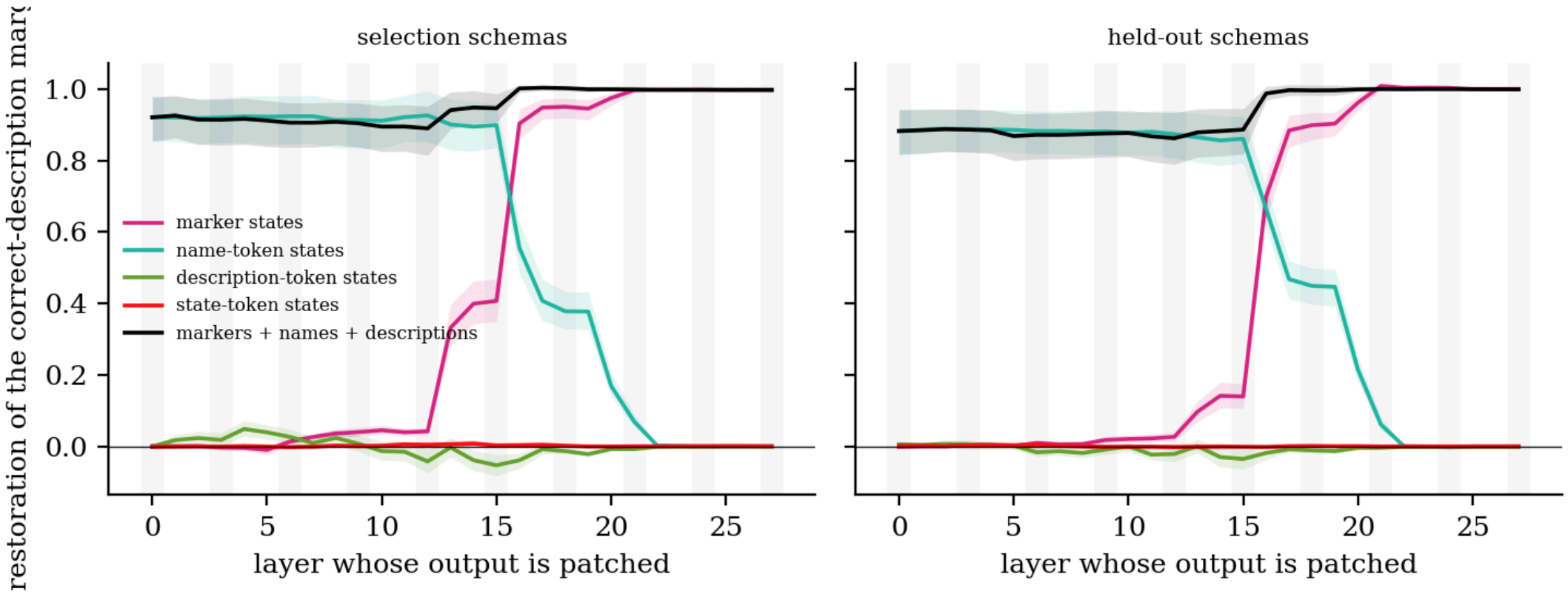

**Figure 16.** Residual patching in v1: restoration of the correct-description margin when the clean states of one token role are copied into the swapped run at the output of each layer. Shaded: bootstrap 95% intervals over items.

**Table 9.** Restoration of the correct-description margin when residual states of the original-name run are copied into the swapped-name run at the output of layer L (v1; 1 = complete, 0 = none). n = 160 swap items per row; for names only options whose name has the same token length in both runs are patched.

| Patched states | Schemas | L = 8 | 12 | 13 | 14 | 15 | 16 | 17 | 20 | 24 |
|---|---|---|---|---|---|---|---|---|---|---|
| markers | selection | 0.04 | 0.04 | 0.33 | 0.40 | 0.41 | 0.90 | 0.95 | 0.97 | 1.00 |
| markers | held-out | 0.01 | 0.03 | 0.10 | 0.14 | 0.14 | 0.70 | 0.88 | 0.96 | 1.00 |
| name tokens | selection | 0.91 | 0.93 | 0.90 | 0.89 | 0.90 | 0.56 | 0.41 | 0.17 | -0.00 |
| name tokens | held-out | 0.88 | 0.87 | 0.87 | 0.86 | 0.86 | 0.66 | 0.47 | 0.21 | -0.00 |
| description tokens | selection | 0.02 | -0.04 | -0.00 | -0.04 | -0.05 | -0.04 | -0.01 | -0.01 | 0.00 |
| description tokens | held-out | -0.02 | -0.02 | 0.00 | -0.03 | -0.04 | -0.02 | -0.01 | -0.00 | -0.00 |
| state tokens | selection | 0.00 | 0.00 | 0.01 | 0.01 | 0.00 | 0.00 | 0.00 | -0.00 | 0.00 |
| state tokens | held-out | 0.00 | -0.00 | -0.00 | -0.00 | -0.00 | -0.00 | 0.00 | 0.00 | -0.00 |
| question tokens | selection | 0.00 | -0.00 | -0.00 | -0.00 | 0.00 | -0.00 | -0.00 | -0.00 | -0.00 |
| question tokens | held-out | 0.00 | -0.00 | 0.00 | 0.00 | 0.00 | 0.00 | -0.00 | -0.00 | 0.00 |
| markers, names and descriptions | selection | 0.91 | 0.89 | 0.94 | 0.95 | 0.95 | 1.00 | 1.00 | 1.00 | 1.00 |
| markers, names and descriptions | held-out | 0.87 | 0.86 | 0.88 | 0.88 | 0.89 | 0.99 | 1.00 | 1.00 | 1.00 |

Three results stand out (Figure 16, Table 9). First, the marker states become sufficient in a narrow range of layers: restoration passes 0.5 at layer 16 on the selection schemas and at layer 16 on the held-out schemas, and passes 0.9 at layers 16 and 19; before layer 13 it is below 0.05. Copying the marker states of any later layer completes the repair, so everything the decision needs from the name has reached the marker by layer 16 on the selection schemas and by layer 19 on the held-out schemas. Second, the name-token states carry the corruption up to the same point and not beyond: restoration through name patching is 0.93 at layer 12 and 0.90 at layer 15, then falls to 0.56 at layer 16, 0.17 at layer 20 and about -0.00 at layer 24 (selection schemas; the held-out values are 0.87, 0.86, 0.66, 0.21 and -0.00). Once the markers have read the names, repairing the names no longer helps. Third, the description, state and question tokens restore nothing at any layer (largest absolute restoration 0.05), and patching markers, names and descriptions together restores at least 0.86 at every layer.

Patching and knockouts agree on the window. Restoring clean marker states before layer 13 achieves nothing, because the markers read the corrupted names again in layers 13 to 16; restoring clean name tokens works until the same point for the same reason. After the window closes the markers hold the corruption and the names no longer matter. The knockouts add that reading in layers 0 to 11 is not needed for the shortcut, which is why the first-third prediction in Section 13 failed.

## 9 Linear probes

Probes ask what is linearly available at the marker. For each layer we fit a ridge classifier on standardised marker states of the swapped inputs of the selection schemas, with one of two labels: the option carries the gold name, or the option has the gold description. The classifiers are tested on the held-out schemas, whose names and descriptions never appeared in the probe training data, so any signal is a cross-schema signal. We also decode name identity within each schema (five folds over items) and report it against a control task in which names are mapped to random classes [26].

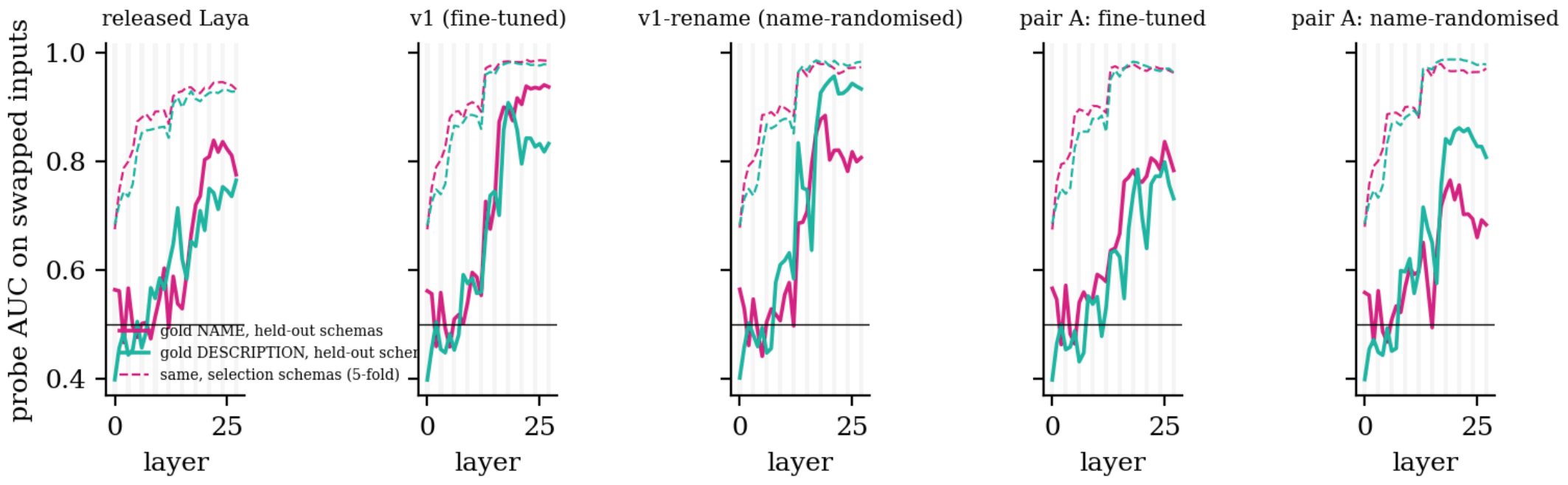

**Figure 17.** Probe AUC by layer on swapped inputs for the option carrying the gold name and the option with the gold description. Solid: held-out schemas; dashed: selection schemas (five folds).

**Table 10.** Ridge probes on marker states of swapped inputs, trained on the selection schemas and tested on the held-out schemas. ‘Name first’ is the first layer at which the AUC for the option carrying the gold NAME exceeds the AUC for the option with the gold DESCRIPTION by 0.1, counted only from the first layer at which the larger of the two AUCs reaches 0.65 (earlier AUCs below 0.5 reflect the schema shift); ‘description first’ is the reverse. Last column: first layer at which name identity is decoded with at least 90% accuracy (mean of four schemas).

| Model | AUC (name) there | Name first (layer) | Description first (layer) | Maximum AUC (name / description) | Name identity 90% (layer) |
|---|---|---|---|---|---|
| released Laya | 0.80 | 20 | 14 | 0.84 / 0.76 | 0 |
| v1 (fine-tuned) | 0.87 | 16 | never | 0.94 / 0.91 | 0 |
| v1-rename (name-randomised) | 0.79 | 16 | 13 | 0.88 / 0.96 | 0 |
| pair A: fine-tuned | 0.76 | 16 | never | 0.84 / 0.80 | 0 |
| pair A: name-randomised | n/a | never | 14 | 0.76 / 0.86 | 0 |

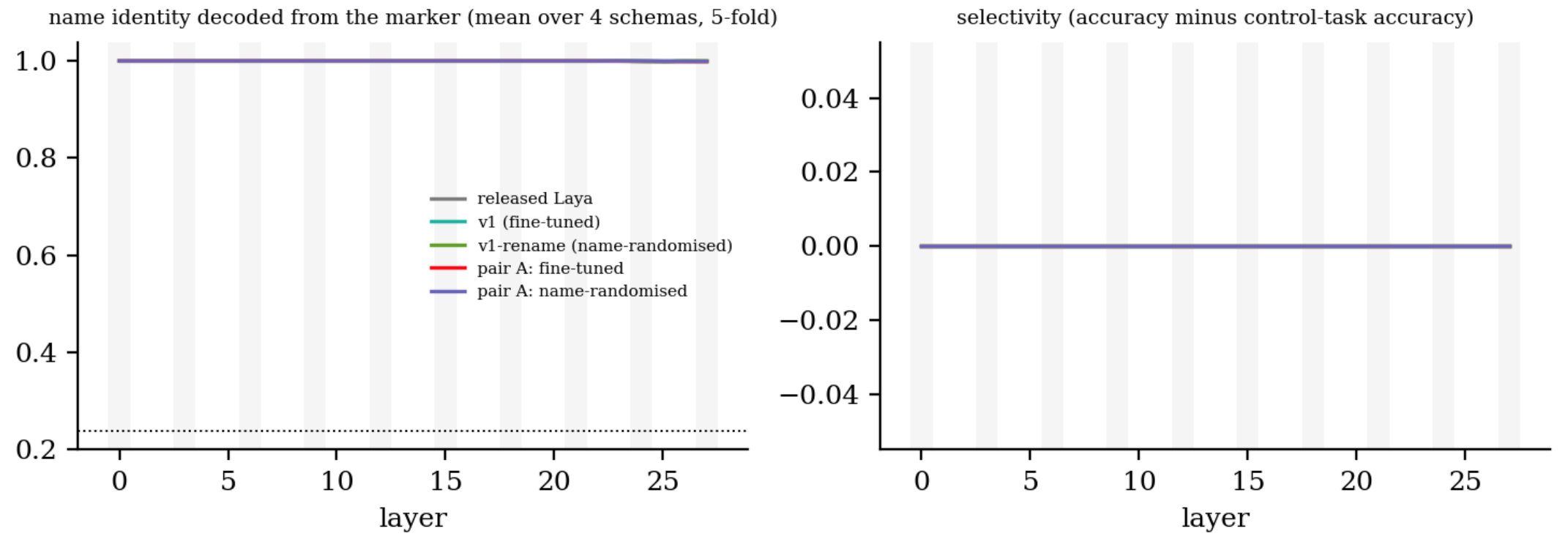

**Figure 18.** Name identity decoded from the marker state (left) and its selectivity over a control task (right).

released Laya: name identity reaches 90% decodability at layer 0; the name-gold signal leads the description-gold signal by 0.1 AUC from layer 20, and the description-gold signal leads from layer 14.

v1 (fine-tuned): name identity reaches 90% decodability at layer 0; the name-gold signal leads the description-gold signal by 0.1 AUC from layer 16, and the description-gold signal never leads.

v1-rename (name-randomised): name identity reaches 90% decodability at layer 0; the name-gold signal leads the description-gold signal by 0.1 AUC from layer 16, and the description-gold signal leads from layer 13.

pair A: fine-tuned: name identity reaches 90% decodability at layer 0; the name-gold signal leads the description-gold signal by 0.1 AUC from layer 16, and the description-gold signal never leads.

pair A: name-randomised: name identity reaches 90% decodability at layer 0; the name-gold signal never leads the description-gold signal by 0.1 AUC, and the description-gold signal leads from layer 14.

**Table 11.** Removing a probe direction from every marker state in four layers (v1). Held-out swap shortcut rate (n = 160; 75.0% without edit) after removing the name-gold direction, the description-gold direction and a random direction, and original-name accuracy after removing the name-gold direction.

| Layers edited | Name-gold removed | Description-gold removed | Random removed | Original names right (name-gold removed) |
|---|---|---|---|---|
| 10-13 | 75.0% [67.8, 81.1] | 75.0% [67.8, 81.1] | 75.0% [67.8, 81.1] | 90.3% [86.6, 93.1] |
| 12-15 | 75.0% [67.8, 81.1] | 75.0% [67.8, 81.1] | 75.0% [67.8, 81.1] | 90.3% [86.6, 93.1] |
| 14-17 | 75.0% [67.8, 81.1] | 75.0% [67.8, 81.1] | 75.6% [68.4, 81.6] | 90.3% [86.6, 93.1] |
| 16-19 | 75.0% [67.8, 81.1] | 75.0% [67.8, 81.1] | 75.0% [67.8, 81.1] | 90.3% [86.6, 93.1] |
| 18-21 | 75.0% [67.8, 81.1] | 75.0% [67.8, 81.1] | 75.0% [67.8, 81.1] | 90.3% [86.6, 93.1] |
| 20-23 | 75.0% [67.8, 81.1] | 75.0% [67.8, 81.1] | 75.0% [67.8, 81.1] | 90.3% [86.6, 93.1] |

Table 11 gives the causal check of the two cross-schema directions in v1. The same edit in released Laya (held-out shortcut 64.4% without edit) lowers the shortcut by at most 0.0 points; v1-rename (name-randomised) (held-out shortcut 14.4% without edit) lowers the shortcut by at most 0.0 points; pair A: fine-tuned (held-out shortcut 66.9% without edit) lowers the shortcut by at most 0.0 points; pair A: name-randomised (held-out shortcut 25.0% without edit) lowers the shortcut by at most 0.0 points. In v1 removing the name-gold direction lowers the held-out shortcut by at most 0.0 points. The edit itself is weak: over the six layer windows it changes the name gap of v1 by a mean of 0.11 logits for the name-gold direction and 0.20 for a random direction (the name gap has a standard deviation of 18.0 across items), probably because the ridge directions lie in low-variance parts of the marker state. The test therefore cannot say whether the name travels along the probe direction; we draw no conclusion about the channel, and a subspace erasure would be the stronger test.

## 10 The same wiring in the released, fine-tuned and name-randomised models?

The knockout of Section 6 was repeated on 5 checkpoints. Table 12 and Figure 11 give the result for the marker-to-name and marker-to-description edges.

**Table 12.** Marker-to-name and marker-to-description knockouts in the five checkpoints (swap items, n = 320; Wilson 95% intervals). 'Best window' is the four-layer window with the lowest shortcut rate.

| Model | No knockout | Name edges blocked, layers 14-17 | Best window for name edges | Name edges, all layers | Description edges blocked, layers 14-17 | Description edges, all layers | Original names right, no knockout | Original names right, name edges blocked 14-17 |
|---|---|---|---|---|---|---|---|---|
| released Laya | 54.1% | 28.1% [23.5, 33.3] | 14-17: 28.1% | 10.9% [8.0, 14.8] | 63.4% [58.0, 68.5] | 64.1% [58.7, 69.1] | 72.2% | 71.2% [66.1, 75.9] |
| v1 (fine-tuned) | 66.9% | 16.2% [12.6, 20.7] | 16-19: 15.0% | 3.8% [2.2, 6.4] | 86.2% [82.0, 89.6] | 88.1% [84.1, 91.2] | 90.3% | 90.0% [86.2, 92.8] |
| v1-rename (name-randomised) | 11.2% | 4.7% [2.9, 7.6] | 14-17: 4.7% | 4.7% [2.9, 7.6] | 55.9% [50.5, 61.3] | 68.8% [63.5, 73.6] | 90.3% | 89.7% [85.9, 92.6] |
| pair A: fine-tuned | 64.4% | 17.5% [13.7, 22.0] | 14-17: 17.5% | 7.5% [5.1, 10.9] | 75.6% [70.6, 80.0] | 76.6% [71.6, 80.9] | 82.2% | 83.4% [79.0, 87.1] |
| pair A: name-randomised | 15.9% | 8.4% [5.9, 12.0] | 14-17: 8.4% | 8.8% [6.1, 12.4] | 58.1% [52.7, 63.4] | 68.8% [63.5, 73.6] | 82.2% | 82.2% [77.6, 86.0] |

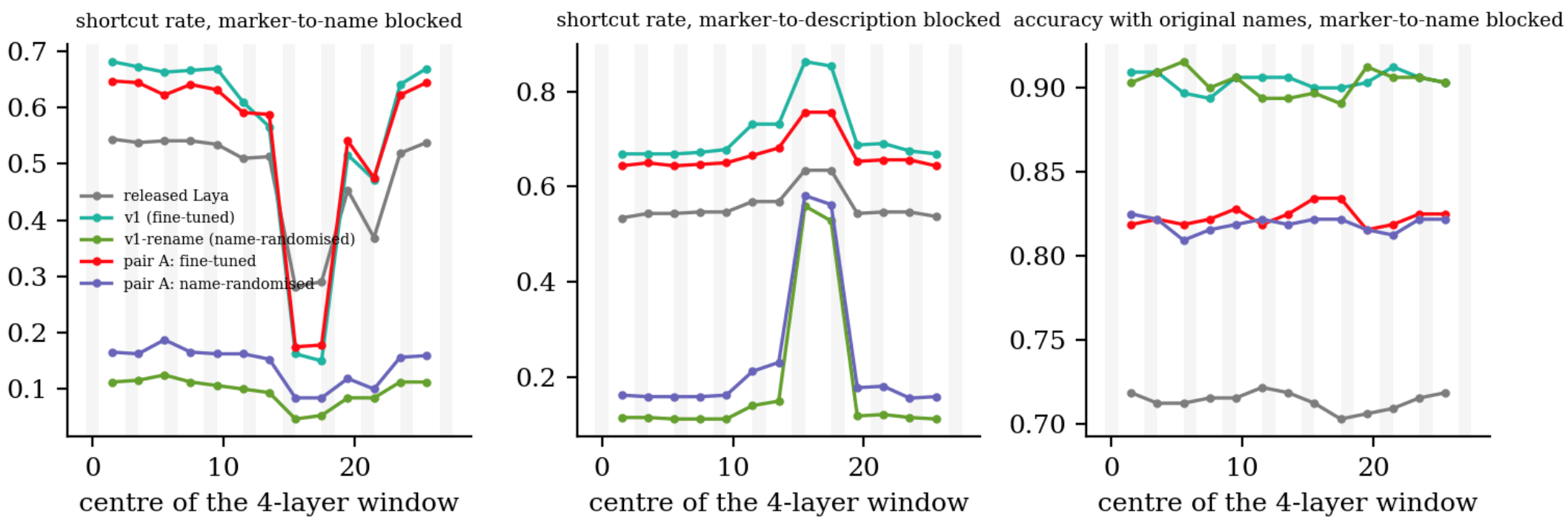

**Figure 11.** Marker-to-name knockouts (left), marker-to-description knockouts (centre) and original-name accuracy under the name knockout (right), by window, for the checkpoints that were run (swap items, n = 320).

The window is similar in all checkpoints: the best four-layer window is layers 14 to 17 in 4 of 5 and a window shifted by two layers in the rest, and blocking the name edges in every layer leaves at most 10.9% of swap items on the name. Blocking the same edges in layers 14 to 17 lowers the rate in every checkpoint (released Laya 54.1% to 28.1%; v1 (fine-tuned) 66.9% to 16.2%; v1-rename (name-randomised) 11.2% to 4.7%; pair A: fine-tuned 64.4% to 17.5%; pair A: name-randomised 15.9% to 8.4%). The released checkpoint keeps the largest remainder after the four-layer edit, so a wider window is needed there. In the two name-randomised checkpoints the rate without knockout is already low, but blocking the marker-to-description edges raises it to a level comparable to the no-knockout rate of the plain checkpoints: the name channel is still wired in and the description channel outvotes it. In these two checkpoints, then, name randomisation added a stronger competitor and did not remove the read.

v1-rename (name-randomised): the shortcut rate without knockout is 11.2%. With the marker-to-description edges blocked in layers 14 to 17 it becomes 55.9%, and with them blocked in all layers 68.8%. The model still reads the names in the same layers and the description channel outvotes them: removing the description channel brings the shortcut back.

pair A: name-randomised: the shortcut rate without knockout is 15.9%. With the marker-to-description edges blocked in layers 14 to 17 it becomes 58.1%, and with them blocked in all layers 68.8%. The model still reads the names in the same layers and the description channel outvotes them: removing the description channel brings the shortcut back.

**Table 13.** First layer at which patching clean marker states restores 50% and 90% of the margin (selection and held-out schemas).

| Model | 50%, selection | 50%, held-out | 90%, selection | 90%, held-out |
|---|---|---|---|---|
| released Laya | 16 | 17 | 20 | 21 |
| v1 (fine-tuned) | 16 | 16 | 16 | 19 |
| v1-rename (name-randomised) | 13 | 16 | 16 | 20 |
| pair A: fine-tuned | 16 | 16 | 17 | 20 |
| pair A: name-randomised | 13 | 16 | 13 | 17 |

Residual patching gives Table 13. In every checkpoint the marker states restore half of the margin between layers 13 and 17, inside or just after the knockout window, and nowhere before layer 13 in v1.

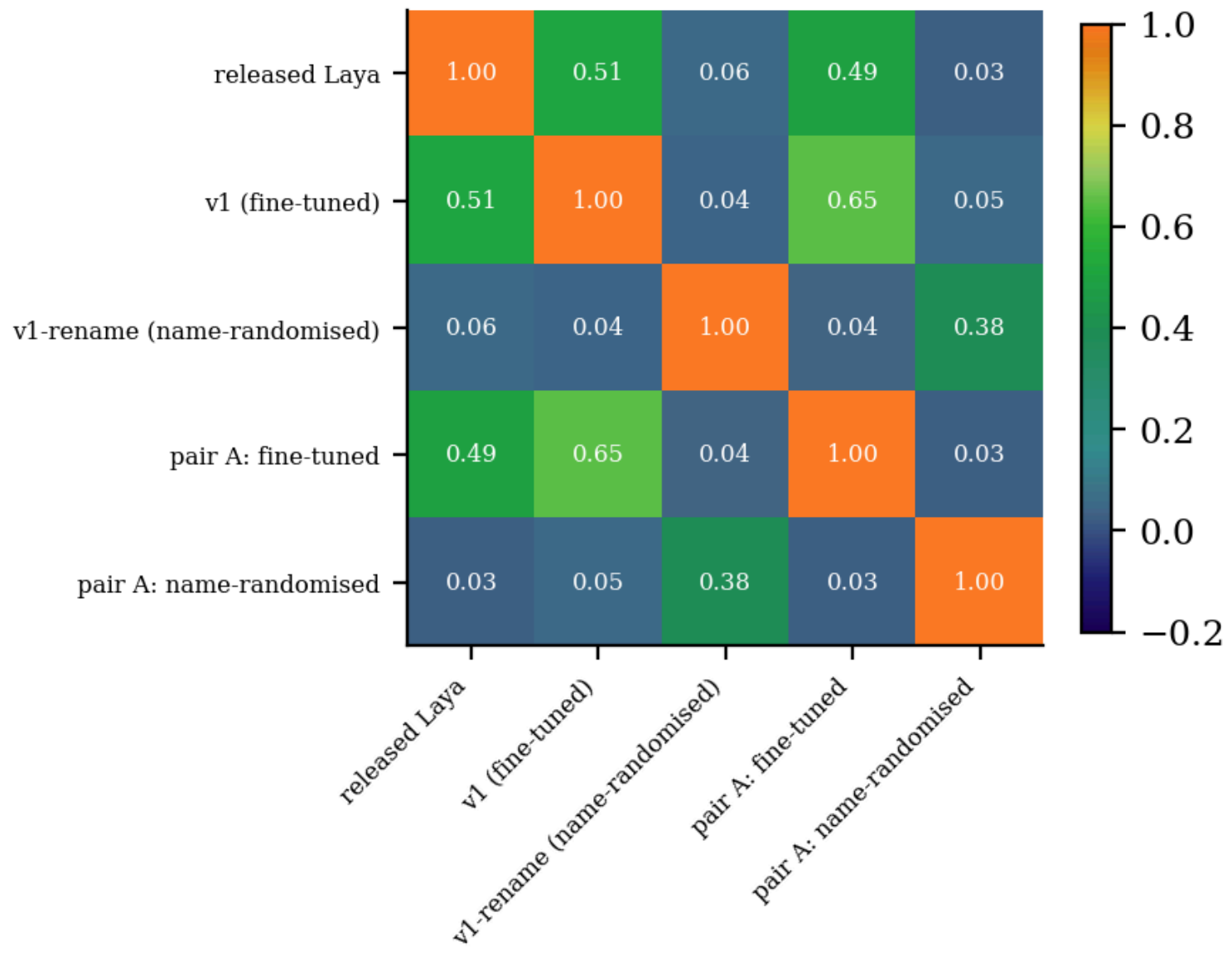

**Figure 23.** Spearman correlation between the per-head effect vectors (drop of the name gap under mean-ablation, 448 heads) of the checkpoints that were swept.

The per-head effect vectors correlate across checkpoints as follows with v1: released Laya 0.51, v1-rename (name-randomised) 0.04, pair A: fine-tuned 0.65, pair A: name-randomised 0.05 (Figure 23). The heads that matter are therefore partly shared between the plain checkpoints (largest correlation 0.65, smallest 0.04). The name-randomised checkpoints have a small name effect to rank, which lowers their correlation by construction.

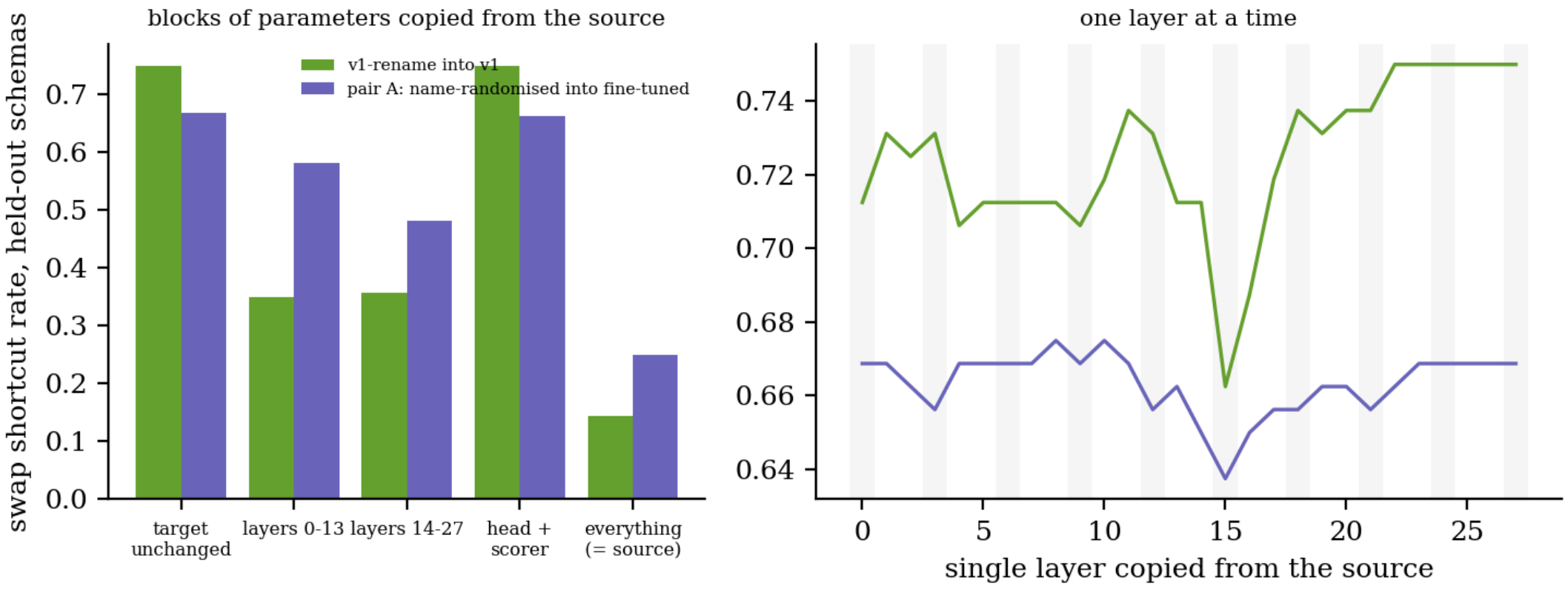

**Figure 19.** Layer grafting: parameters of a source checkpoint copied into a target checkpoint. Left: swap shortcut rate on the held-out schemas after copying a block. Right: after copying one layer.

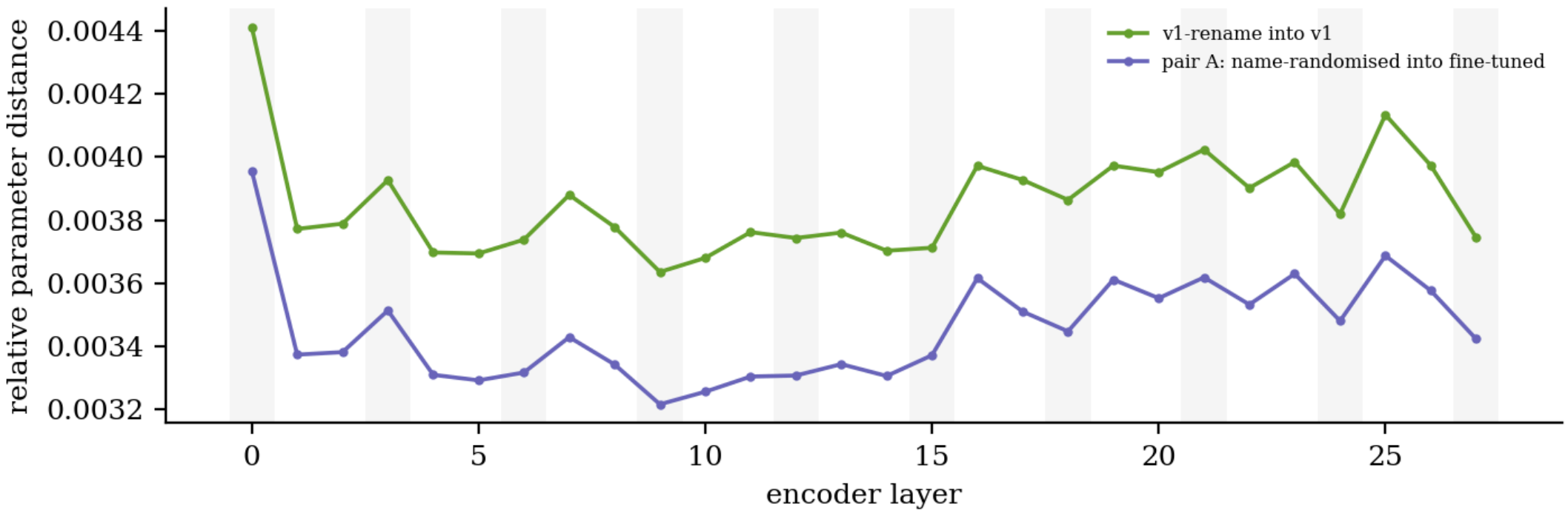

**Figure 20.** Relative parameter distance between source and target per encoder layer.

**Table 14.** Swap shortcut rate on the held-out schemas after grafting blocks of parameters (n = 160).

| Graft | Target as is | Layers 0-13 copied | Layers 14-27 copied | Head and scorer copied | Everything copied |
|---|---|---|---|---|---|
| v1-rename into v1 | 75.0% [67.8, 81.1] | 35.0% [28.0, 42.7] | 35.6% [28.6, 43.3] | 75.0% [67.8, 81.1] | 14.4% [9.8, 20.6] |
| pair A: name-randomised into fine-tuned | 66.9% [59.3, 73.7] | 58.1% [50.4, 65.5] | 48.1% [40.5, 55.8] | 66.2% [58.6, 73.1] | 25.0% [18.9, 32.2] |

Table 14 and Figure 19 give the grafting results.

v1-rename into v1: the held-out shortcut is 75.0% in the target and 14.4% with every parameter copied. Copying layers 0 to 13 gives 35.0%, layers 14 to 27 35.6%, and the head and scorer 75.0%; as shares of the full reduction these are 66%, 65% and 0%. The two encoder halves were trained together, so this is a decomposition of what was copied and not a localisation of where the fix lives.

pair A: name-randomised into fine-tuned: the held-out shortcut is 66.9% in the target and 25.0% with every parameter copied. Copying layers 0 to 13 gives 58.1%, layers 14 to 27 48.1%, and the head and scorer 66.2%; as shares of the full reduction these are 21%, 45% and 1%. The two encoder halves were trained together, so this is a decomposition of what was copied and not a localisation of where the fix lives.

## 11 The option head

After the encoder, two transformer layers without positional encoding mix all token states before the scorer reads each marker. Is the decision made there? Figure 21 shows where marker queries look in these two layers: 29.0% of the mass of the first head layer goes to markers, 19.2% to the state and 28.9% to the question; the second head layer gives 33.3% to markers. Table 15 removes both layers.

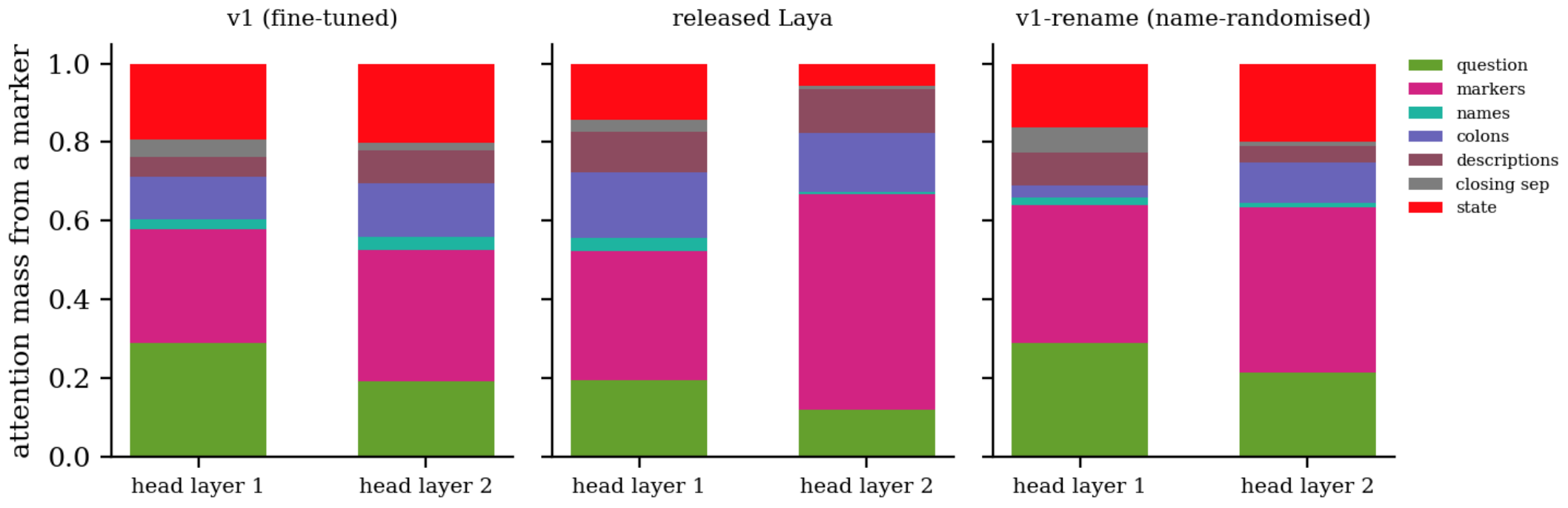

**Figure 21.** Attention mass from marker tokens in the two option-head layers (mean over heads and original-name items).

**Table 15.** Accuracy with the original names on the shared sample (n = 320) when the two option-head layers are skipped (the encoder output at each marker goes straight to the scorer), against accuracy with the head. For v1 the comparison is on the same 320 items; for the other checkpoints the intact-head figure comes from the full 951-item pool and is only indicative.

| Model | Head skipped | With head |
|---|---|---|
| released Laya | 71.2% | 69.9% (full pool) |
| v1 (fine-tuned) | 89.4% | 90.3% |
| v1-rename (name-randomised) | 89.1% | 87.5% (full pool) |
| pair A: fine-tuned | 77.8% | 81.5% (full pool) |
| pair A: name-randomised | 80.6% | 84.1% (full pool) |

Skipping the head leaves v1 at 89.4% against 90.3% with the head, a change of -0.9 points on 320 items. The scorer can read the decision straight from the encoder output at the marker. The other checkpoints behave alike, with the caveat in the table caption.

## 12 A training-free name-blind mode

If the name is read through a few identifiable edges, those edges can be cut at inference time. We evaluate the edit of Section 6 on the full pools: no retraining, no new data, only a mask on attention from each marker to its own name and colon tokens in a window of layers.

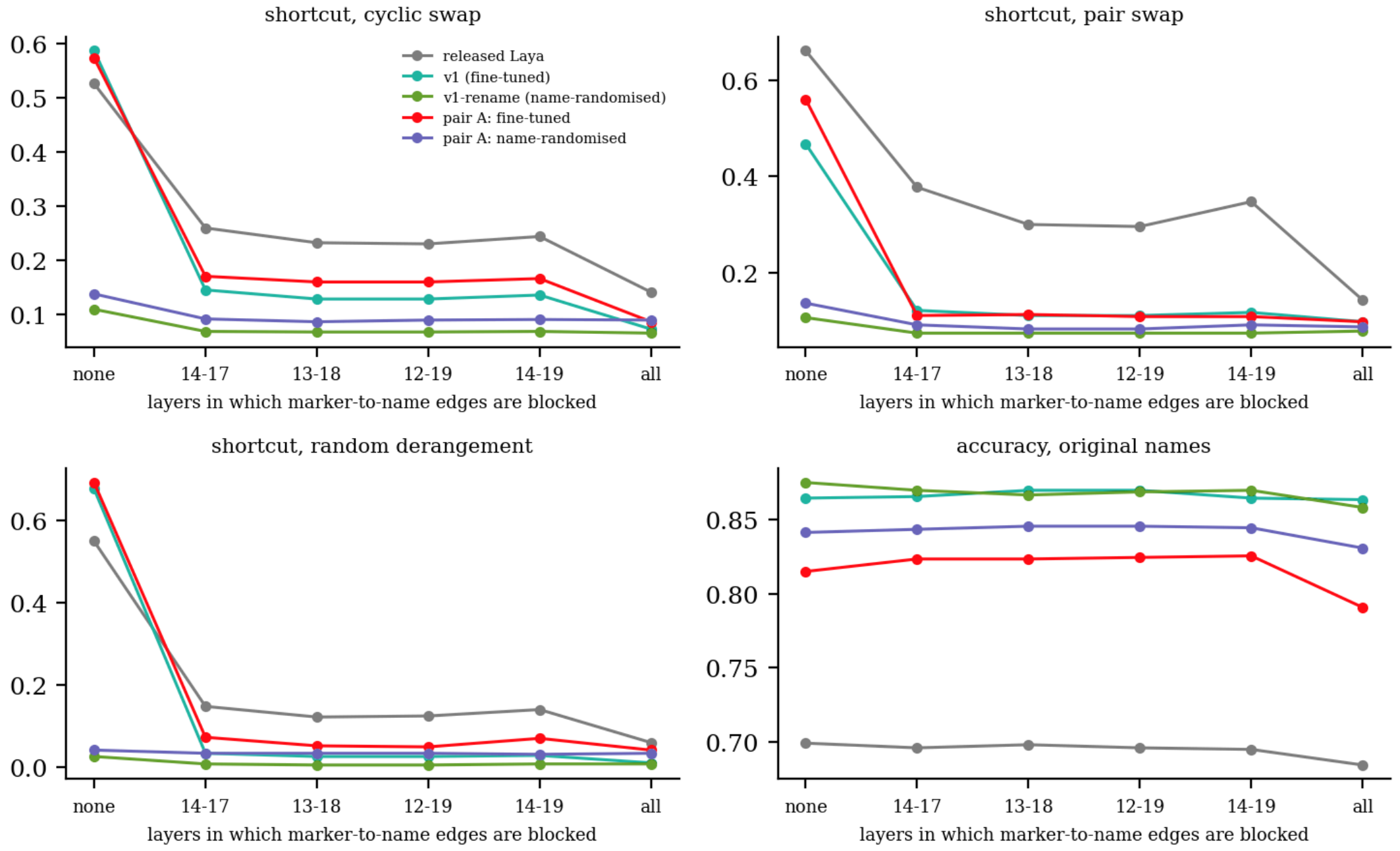

**Figure 22.** Marker-to-name edges blocked at inference time, by layer window, on the full conflict pools.

**Table 16.** Blocking the marker-to-name edges at inference time on the full pools (orig 951, swap 951, pair 466, derangement 385, no-description 951 items). Percent; Wilson 95% intervals.

| Model | Edit | Original names right | Swap: name followed | Pair: name followed | Derangement: name followed | No descriptions: right |
|---|---|---|---|---|---|---|
| released Laya | no edit | 69.9% [66.9, 72.8] | 52.7% [49.5, 55.8] | 66.1% [61.7, 70.2] | 55.1% [50.1, 60.0] | 69.6% [66.6, 72.5] |
| released Laya | layers 14-17 | 69.6% [66.6, 72.5] | 26.0% [23.3, 28.9] | 37.8% [33.5, 42.3] | 14.8% [11.6, 18.7] | 63.2% [60.1, 66.2] |
| released Laya | layers 12-19 | 69.6% [66.6, 72.5] | 23.0% [20.5, 25.8] | 29.6% [25.6, 33.9] | 12.5% [9.5, 16.1] | 56.5% [53.3, 59.6] |
| released Laya | all layers | 68.5% [65.4, 71.3] | 14.1% [12.0, 16.4] | 14.4% [11.5, 17.9] | 6.0% [4.0, 8.8] | 17.0% [14.8, 19.6] |
| v1 (fine-tuned) | no edit | 86.4% [84.1, 88.5] | 58.8% [55.6, 61.9] | 46.8% [42.3, 51.3] | 67.8% [63.0, 72.3] | 83.8% [81.3, 86.0] |
| v1 (fine-tuned) | layers 14-17 | 86.5% [84.2, 88.6] | 14.5% [12.4, 16.9] | 12.2% [9.6, 15.5] | 3.4% [2.0, 5.7] | 61.5% [58.4, 64.6] |
| v1 (fine-tuned) | layers 12-19 | 87.0% [84.7, 89.0] | 12.8% [10.9, 15.1] | 11.2% [8.6, 14.3] | 2.6% [1.4, 4.7] | 46.9% [43.7, 50.1] |
| v1 (fine-tuned) | all layers | 86.3% [84.0, 88.4] | 7.3% [5.8, 9.1] | 9.9% [7.5, 12.9] | 1.0% [0.4, 2.6] | 12.7% [10.8, 15.0] |
| v1-rename (name-randomised) | no edit | 87.5% [85.2, 89.4] | 10.9% [9.1, 13.1] | 10.7% [8.2, 13.9] | 2.6% [1.4, 4.7] | 83.0% [80.4, 85.2] |
| v1-rename (name-randomised) | layers 14-17 | 87.0% [84.7, 89.0] | 6.8% [5.4, 8.6] | 7.5% [5.4, 10.3] | 0.8% [0.3, 2.3] | 24.8% [22.2, 27.7] |
| v1-rename (name-randomised) | layers 12-19 | 86.9% [84.6, 88.9] | 6.7% [5.3, 8.5] | 7.5% [5.4, 10.3] | 0.5% [0.1, 1.9] | 17.1% [14.9, 19.7] |
| v1-rename (name-randomised) | all layers | 85.8% [83.4, 87.9] | 6.5% [5.1, 8.3] | 7.9% [5.8, 10.8] | 0.8% [0.3, 2.3] | 10.5% [8.7, 12.6] |
| pair A: fine-tuned | no edit | 81.5% [78.9, 83.8] | 57.4% [54.2, 60.5] | 56.0% [51.5, 60.4] | 69.4% [64.6, 73.7] | 80.4% [77.8, 82.8] |
| pair A: fine-tuned | layers 14-17 | 82.3% [79.8, 84.6] | 17.0% [14.8, 19.6] | 11.2% [8.6, 14.3] | 7.3% [5.1, 10.3] | 58.0% [54.9, 61.1] |
| pair A: fine-tuned | layers 12-19 | 82.4% [79.9, 84.7] | 16.0% [13.8, 18.4] | 10.9% [8.4, 14.1] | 4.9% [3.2, 7.6] | 44.4% [41.2, 47.5] |
| pair A: fine-tuned | all layers | 79.1% [76.4, 81.5] | 8.6% [7.0, 10.6] | 9.9% [7.5, 12.9] | 4.2% [2.6, 6.6] | 16.4% [14.2, 18.9] |
| pair A: name-randomised | no edit | 84.1% [81.7, 86.3] | 13.8% [11.7, 16.1] | 13.7% [10.9, 17.2] | 4.2% [2.6, 6.6] | 80.7% [78.0, 83.0] |
| pair A: name-randomised | layers 14-17 | 84.3% [81.9, 86.5] | 9.1% [7.5, 11.1] | 9.2% [6.9, 12.2] | 3.4% [2.0, 5.7] | 29.4% [26.6, 32.4] |
| pair A: name-randomised | layers 12-19 | 84.5% [82.1, 86.7] | 8.9% [7.3, 10.9] | 8.4% [6.2, 11.2] | 3.4% [2.0, 5.7] | 22.9% [20.4, 25.7] |
| pair A: name-randomised | all layers | 83.1% [80.6, 85.3] | 8.9% [7.3, 10.9] | 8.8% [6.6, 11.7] | 3.4% [2.0, 5.7] | 11.0% [9.2, 13.2] |

Table 16 and Figure 22 give the result. This is a diagnostic and not a recommended deployment: it removes the name channel for every question, including those where the name carries information the description lacks, and it is tested here only on four synthetic schemas.

## 13 Predictions against outcomes

Table 17 lists the twelve predictions (idea 10 of the pre-run list is the option head, idea 12 the description deletion) with the observed values. We count a prediction as contradicted if its stated inequality fails on the held-out schemas (or, where no held-out split exists, on the shared sample). Of the 12 predictions that could be scored, 9 were contradicted.

**Table 17.** Predictions written before the runs and what happened. 'Not run' means the experiment did not complete (Section 14).

| # | Prediction | Observed | Outcome |
|---|---|---|---|
| 1 | Marker-to-name edges are read early: blocking in layers 0-7 cuts the shortcut by at least 15 points, in layers 16-27 by at most 5. | layers 0-7: at most 0.6 points; layers 16-27: up to 51.9 points | contradicted |
| 2 | Decision assembly is late: blocking marker-to-state edges lowers original-name accuracy more than any option-side knockout. | all layers: accuracy 90.3% against 90.3% without knockout | contradicted |
| 3 | Name reading is sparse: the top 10 heads hold at least half of the summed effect, and ablating them cuts the held-out shortcut by at least 20 points with at most 10 points of accuracy loss. | top 10 hold 29%; held-out swap shortcut 75.0% to 66.9%; accuracy 90.3% to 87.5% | contradicted |
| 4 | Marker states restore the decision between layers 10 and 16 and exceed 90% by layer 24. | 50% at layers 16 and 16, 90% at layers 16 and 19 | confirmed |
| 5 | Blocking name-to-description edges in layers 0-7 removes at least a third of the effect of blocking marker-to-name edges. | at most 0.6 points | contradicted |
| 6 | Name identity is decodable (90%) by layer 6 in every checkpoint; in v1 the name-gold probe leads by layer 12, in v1-rename the description-gold probe leads. | released Laya: 90% at layer 0, name leads from 20, description leads from 14; v1 (fine-tuned): 90% at layer 0, name leads from 16, description leads from never; v1-rename (name-randomised): 90% at layer 0, name leads from 16, description leads from 13; pair A: fine-tuned: 90% at layer 0, name leads from 16, description leads from never; pair A: name-randomised: 90% at layer 0, name leads from never, description leads from 14 | contradicted |
| 7 | Name-randomised fine-tuning leaves name identity decodable: within 5 points of v1 at every layer. | v1-rename (name-randomised): largest gap 0.1 points; pair A: name-randomised: largest gap 0.1 points | confirmed |
| 8 | Copying layers 14-27 (plus head) from the name-randomised model restores at least 60% of its shortcut reduction; layers 0-13 alone less than 30%. | v1-rename (name-randomised) into v1 (fine-tuned): layers 14-27 65%, layers 0-13 66%, head and scorer 0%; pair A: name-randomised into pair A: fine-tuned: layers 14-27 45%, layers 0-13 21%, head and scorer 1% | contradicted |
| 9 | Head-effect vectors of the released checkpoint and v1 correlate at least 0.6 (Spearman). | 0.51 | contradicted |
| 10 | At least 40% of marker attention in the head layers goes to other markers, and removing both head layers costs more than 30 points. | 29% and 33% on markers; removal changes accuracy from 90.3% to 89.4% | contradicted |

| # | Prediction | Observed | Outcome |
|---|---|---|---|
| 11 | Erasing the name-gold direction cuts the held-out swap shortcut by at least 20 points; a random direction moves it by less than 5. | best layers 10-13: 75.0% to 75.0%; random direction at most 0.6 points (the name-gold edit moves the name gap less than a random one does, so the test is weak; Section 9) | contradicted |
| 12 | Deleting descriptions changes accuracy by less than 5 points. | v1: 2.6 points [0.8, 4.5] | confirmed |

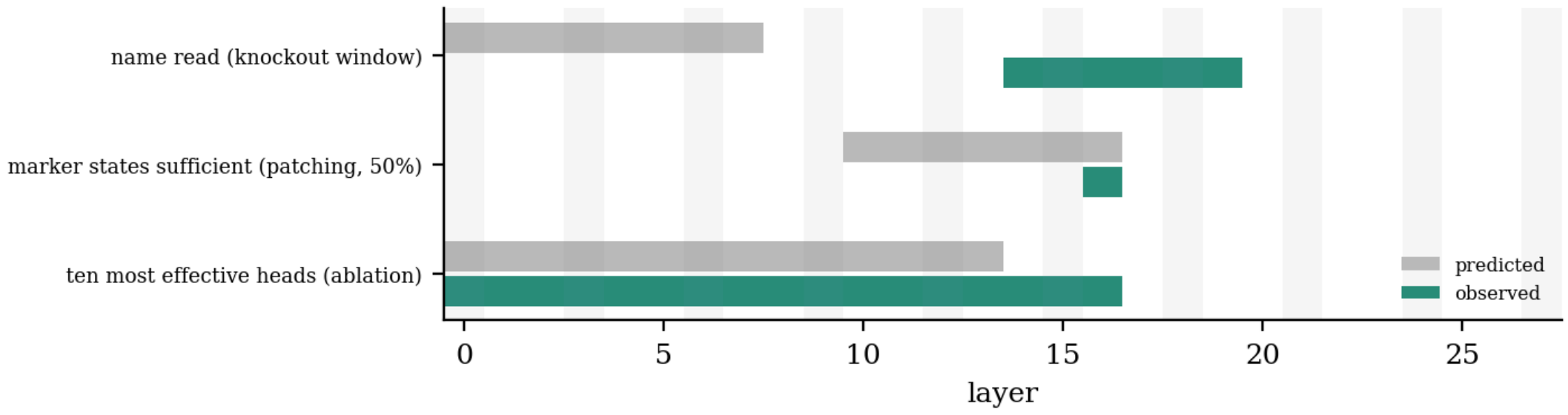

**Figure 24.** Layers predicted before the runs (grey) and layers observed (blue) for three localisation experiments in v1.

## 14 Limitations and threats to validity

- Synthetic items and LLM labels. All 951 conflict questions were written by one generator model (Qwen3.8-27B) and labelled by a blind second pass of the same model; an independent labeller agrees on 94 to 99% of the items per set. The test schemas are four short, familiar sets of options with a mean input length of 101 tokens. Real tickets, longer texts and rubric-style descriptions may route differently, and in a longer input the local-attention layers would not see the whole sequence.
- One architecture family. Everything here concerns Laya and its fine-tunes (ModernBERT-large with a two-layer head). We examined the multilingual and typed-decisions releases only behaviourally (Table 3). A different encoder, or a model that renders options without the name [3], can have a different circuit.
- Interventions leave the data distribution. A knockout, a mean ablation and a patch each create activations the model never saw in training. We use controls of other kinds in the same layers and one zero-ablation check, but we do not know how a different ablation would change the layer boundaries by a layer or two. For name patching, only the 86% of option spans whose name has the same token length in both runs are patched.
- No minimal circuit. The name read is redundant across heads and layers: no single head carries it, and the top ten heads carry 29% of the summed head effect. We make no claim about a sparse or complete circuit. The knockouts locate where the edges must exist, and the head ablations show that the implementation is distributed.
- Sample sizes. Each intervention runs on 80 questions per schema (320 swap items, of which 160 held-out; about 60 held-out pair items; 80 held-out derangement items). Rates on held-out subsets have Wilson intervals about 6 to 9 points wide. Head ranking uses 80 swap items, and the one-head effects (largest 8.8 points) are within noise individually.
- What ‘held-out’ means. For pair A the held-out schemas were never seen in fine-tuning. For v1, v1-rename and the released checkpoint they are a selection/test split of components, not of training data (Table 1), so those results show that a component choice transfers across schemas, not that the model generalises to schemas it never saw.
- Probes are correlational. A probe that decodes a property shows that the property is linearly available, not that the model uses it [25, 26]. The erasure test in Section 9 is a causal check for two cross-schema directions only, and the edit was too weak to be informative.
- Pre-registration is informal. The predictions are in a file written before the runs (in the code repository), but they were not time-stamped by a third party. The author could not review them in advance.

- Numerics. The harness runs in fp32 with eager attention; Laya's own agent runs bf16 with fused attention. The two agree on every one of the 48 items we compared; we did not compare them on the full pool, and a rare flip in bf16 would not change a conclusion of this paper.
- Coverage. Not every checkpoint was given every experiment. Missing: nothing. Grafting ran for v1-rename (name-randomised) into v1 (fine-tuned); pair A: name-randomised into pair A: fine-tuned. The single-edge knockout (marker to its own name tokens only, as opposed to name and colon tokens) ran for v1 only. Comparisons across checkpoints in Sections 8 to 12 are limited to the checkpoints named in each table.

## 15 Conclusion

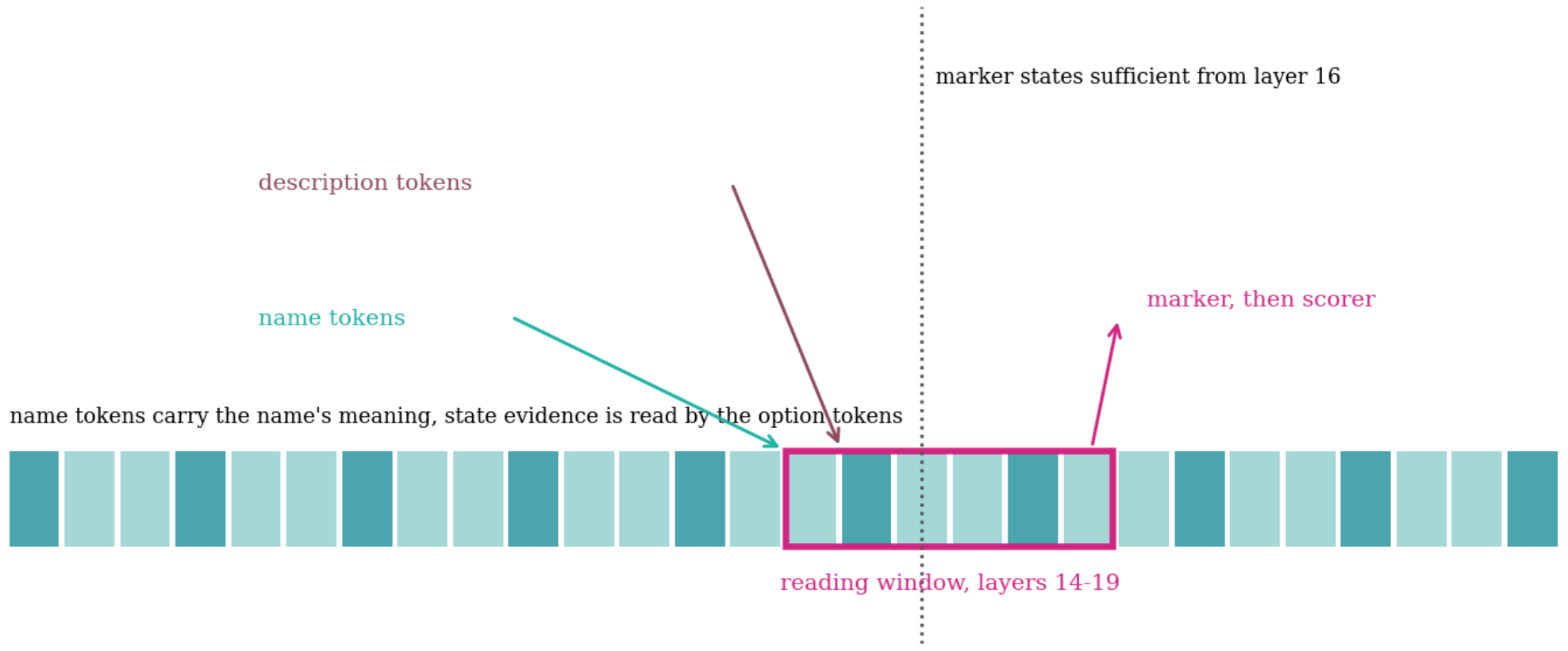

**Figure 25.** Where v1 reads the option name: option tokens carry name and description information (including the text) up to the reading window; markers read them there; the scorer reads the marker.

In v1, the name shortcut of Laya runs through attention edges from each option's marker to its own name tokens in layers 14 to 19 of 28. Cutting those edges removes most of the shortcut without touching accuracy with the original names, and cutting the marker's edges to the option's description in the same layers strengthens it. The name does not travel through the description tokens, and the markers do not need the state directly. No single head carries the read, and head-level attention mass does not predict which heads matter. Marker states are sufficient for the decision from layer 16 on.

Of the predictions we could score, 9 of 12 failed. The failures concern location, the sparsity of the heads, the role of the option head and the binding between names and descriptions; the patching prediction and the description-deletion prediction held. The result applies to Laya on four synthetic schemas; a model whose rendering leaves out the name [3] has no such edge to cut.

## 16 Reproducibility and notes

Everything in this paper is recomputed at build time from cached result files in `results/a17_*.json` (per-item arrays for every intervention, so intervals can be recomputed) and from the data files in `data/`. The experiment harness (`a17_core.py`) and the experiments (`a17_exp.py`) run on one RTX 4090 through a slot guard that limits the machine to two concurrent GPU jobs; every experiment writes a resumable result file. The checkpoints are the released `convaiinnovations/laya` and the local fine-tunes described in Section 3.3; fine-tunes were not retrained for this study. No new labelled data were created, no API key was used, and no checkpoint was written or kept by this study.

The text was drafted and edited with LLM assistance; all numbers are recomputed from the stored results by the build scripts and were checked by the authors.

Software: torch 2.11 with CUDA 12.8, transformers 5.20, laya 0.4.26, numpy; all experiments are fp32. We replace `eager_attention_forward` of the ModernBERT implementation by a function that records attention, blocks edges and ablates heads, and use forward hooks on the 28 encoder layers for patching, capture and edits.

## 17 Cited works

Each of the 36 works cited was read at its arXiv abstract page between 20 November 2026 and the date of this version; four of them (Azizi et al., Geva et al., Wang et al. on label anchors and Wiegreffe et al. on multiple-choice answering) were also read in their method sections. Typed-decision papers were also found through Hugging Face Papers searches, which we list in Appendix C.

## Appendix A  Per-model numbers

**Table 2.** Wilson 95% intervals for the five main checkpoints.

| Model | orig names | swap: description right | swap: name followed | pair: name followed | derangement: name followed |
|---|---|---|---|---|---|
| released Laya | 69.9% [66.9, 72.8] | 29.8% [26.9, 32.7] | 52.7% [49.5, 55.8] | 66.1% [61.7, 70.2] | 55.1% [50.1, 60.0] |
| v1 (fine-tuned) | 86.4% [84.1, 88.5] | 33.1% [30.2, 36.2] | 58.8% [55.6, 61.9] | 46.8% [42.3, 51.3] | 67.8% [63.0, 72.3] |
| v1-rename (name-randomised) | 87.5% [85.2, 89.4] | 82.0% [79.5, 84.3] | 10.9% [9.1, 13.1] | 10.7% [8.2, 13.9] | 2.6% [1.4, 4.7] |
| pair A: fine-tuned | 81.5% [78.9, 83.8] | 31.1% [28.3, 34.1] | 57.4% [54.2, 60.5] | 56.0% [51.5, 60.4] | 69.4% [64.6, 73.7] |
| pair A: name-randomised | 84.1% [81.7, 86.3] | 78.8% [76.0, 81.2] | 13.8% [11.7, 16.1] | 13.7% [10.9, 17.2] | 4.2% [2.6, 6.6] |

**Table 4.** Mean attention mass from marker tokens by token role, v1, original and swapped inputs (mean over heads and layers).

| Role | Original names | Swapped names |
|---|---|---|
| question | 24.9% | 24.8% |
| other marker | 17.7% | 17.2% |
| own marker | 11.0% | 10.7% |
| other name | 2.9% | 3.3% |
| own name | 5.4% | 5.7% |
| other colon | 2.7% | 2.9% |
| own colon | 2.8% | 2.6% |
| other desc | 5.7% | 6.1% |
| own desc | 3.7% | 3.7% |
| closing sep | 5.8% | 5.8% |
| state | 17.3% | 17.2% |

## Appendix B  Heads with the largest effects

**Table 8.** The 20 heads of v1 with the largest drop of the name gap under swap when mean-ablated (layer.head), with the change in shortcut rate and in accuracy with the original names (selection items), and the share of the head's attention that a marker places on its own name tokens.

| Head | Drop of name gap (logits) | Change in shortcut rate (points) | Change in original-name accuracy (points) | Own-name attention |
|---|---|---|---|---|
| 16.3 | 3.81 | +8.8 | +0.0 | 71.8% |
| 15.4 | 3.70 | +8.8 | +0.0 | 0.0% |
| 0.5 | 2.62 | +7.5 | +0.0 | 0.0% |
| 12.8 | 2.37 | +6.2 | +0.0 | 0.4% |
| 13.4 | 2.18 | +1.2 | -2.5 | 51.3% |
| 15.9 | 2.06 | +6.2 | +0.0 | 0.2% |
| 11.11 | 1.86 | +2.5 | +0.0 | 0.4% |
| 11.15 | 1.86 | +1.2 | +0.0 | 0.1% |

| Head | Drop of name gap (logits) | Change in shortcut rate (points) | Change in original-name accuracy (points) | Own-name attention |
|---|---|---|---|---|
| 14.7 | 1.65 | +6.2 | +0.0 | 1.1% |
| 11.8 | 1.35 | +5.0 | +0.0 | 0.6% |
| 0.13 | 1.33 | +1.2 | +0.0 | 0.0% |
| 15.2 | 1.26 | +1.2 | +0.0 | 0.6% |
| 12.7 | 1.25 | +5.0 | +0.0 | 0.6% |
| 8.14 | 1.23 | +2.5 | +0.0 | 1.2% |
| 20.2 | 1.19 | +3.7 | +0.0 | 64.5% |
| 12.3 | 1.18 | -1.3 | +0.0 | 0.6% |
| 14.6 | 1.15 | +2.5 | +0.0 | 2.4% |
| 16.4 | 1.13 | +1.2 | +0.0 | 0.2% |
| 14.14 | 1.11 | +1.2 | +0.0 | 1.6% |
| 21.14 | 1.10 | +2.5 | +0.0 | 73.6% |

## Appendix C Frozen inputs and code

**Table 18.** Result files behind every number (first 16 hex digits of the SHA-256).

| File | Size | SHA-256 |
|---|---|---|
| `results/a17_ablate__a2_ft_ho.json` | 951 KiB | `c16e8786b21d934e` |
| `results/a17_ablate__a2_nr_ho.json` | 976 KiB | `62049c0525fa0ee3` |
| `results/a17_ablate__base.json` | 916 KiB | `26186e997b1c265b` |
| `results/a17_ablate__v1.json` | 955 KiB | `f79f96a85e0c97e8` |
| `results/a17_ablate__v1_rename.json` | 985 KiB | `90816a3dbce22cfe` |
| `results/a17_attn__a2_ft_ho.json` | 743 KiB | `fc4051a4aef68dff` |
| `results/a17_attn__a2_nr_ho.json` | 742 KiB | `ee018a1e6f75bd94` |
| `results/a17_attn__base.json` | 743 KiB | `629e3009a9d9aee9` |
| `results/a17_attn__v1.json` | 742 KiB | `0fda7c13dbe9fcee` |
| `results/a17_attn__v1_rename.json` | 741 KiB | `a2142c93518c400b` |
| `results/a17_behav__a2_ft_ho.json` | 201 KiB | `1b05a8bce6f31cce` |
| `results/a17_behav__a2_nr_ho.json` | 203 KiB | `c8c35abc01a8ecb2` |
| `results/a17_behav__base.json` | 198 KiB | `b1442ab8282e4f65` |
| `results/a17_behav__ml0.json` | 200 KiB | `934e0ed838143b6b` |
| `results/a17_behav__td.json` | 198 KiB | `0152f4f78a3af14f` |
| `results/a17_behav__v1.json` | 201 KiB | `dd4a472bc04d96f9` |
| `results/a17_behav__v1_ren_neutral.json` | 202 KiB | `35091742d1eb8ed6` |
| `results/a17_behav__v1_ren_shift.json` | 203 KiB | `8e803dcb76a1aa7c` |
| `results/a17_behav__v1_rename.json` | 203 KiB | `1c8ec0de24890a22` |
| `results/a17_behav__v1_rename_s1.json` | 203 KiB | `4646ed29916f4b79` |
| `results/a17_behav__v1_rename_s2.json` | 203 KiB | `c6e8f4627a309516` |
| `results/a17_behav__v1_s1.json` | 201 KiB | `ee4342ed386d7718` |
| `results/a17_behav__v1_s2.json` | 202 KiB | `ca9b195d4e450aed` |

| File | Size | SHA-256 |
|---|---|---|
| results/a17_behav__v1_s3.json | 201 KiB | 5514ca23a9739eaf |
| results/a17_graft__a2_ft_ho__from__a2_nr_ho.json | 431 KiB | 69b340cf192e54f6 |
| results/a17_graft__v1__from__v1_rename.json | 433 KiB | 95acd16187020255 |
| results/a17_knock1__v1.json | 1842 KiB | 4451fb488c078c43 |
| results/a17_knock__a2_ft_ho.json | 1251 KiB | b14a63cfe82534fd |
| results/a17_knock__a2_nr_ho.json | 1277 KiB | 28318fe6ee1c5a5b |
| results/a17_knock__base.json | 1197 KiB | 1866521adeb6b6b9 |
| results/a17_knock__v1.json | 1011 KiB | 98d80406162c42bf |
| results/a17_knock__v1_rename.json | 1286 KiB | fe70030a6067a950 |
| results/a17_mitig__a2_ft_ho.json | 456 KiB | 0fee260f4746197b |
| results/a17_mitig__a2_nr_ho.json | 460 KiB | 8623256681147f08 |
| results/a17_mitig__base.json | 436 KiB | 2a3ec5ab794833b5 |
| results/a17_mitig__v1.json | 456 KiB | 28a56acb7d3c5a3e |
| results/a17_mitig__v1_rename.json | 461 KiB | aaf2d1427ba7c819 |
| results/a17_patch__a2_ft_ho.json | 1571 KiB | f472aeef89ee0701 |
| results/a17_patch__a2_nr_ho.json | 1550 KiB | 27e9aa5a12040c7b |
| results/a17_patch__base.json | 1580 KiB | 4349e4a1b69be9ae |
| results/a17_patch__v1.json | 1567 KiB | c632f9d300c543e6 |
| results/a17_patch__v1_rename.json | 1547 KiB | 669940dae5acc075 |
| results/a17_probe__a2_ft_ho.json | 270 KiB | 0096a63cd55478c0 |
| results/a17_probe__a2_nr_ho.json | 278 KiB | bd06727f5d4fcb6c |
| results/a17_probe__base.json | 259 KiB | 3e3cfc54c36a99c1 |
| results/a17_probe__v1.json | 272 KiB | a4397aed9eb1ed37 |
| results/a17_probe__v1_rename.json | 280 KiB | 66f8cb7e2e7666de |

Literature searches. The codename was checked with the Hugging Face Papers search endpoint (query *Deneb-Vritra*, no result). Two further queries (*typed decision model mechanistic interpretability* and *Jev option label attention heads layers*) returned no paper that analyses the inside of a typed-decision model. The 36 cited papers were looked up by arXiv identifier through the arXiv query interface and their abstracts read; the typed-decision papers were first found through the Hugging Face search used in an earlier related-work check.

Code: `a17_core.py` (harness: attention replacement, hooks, item encoding); `a17_exp.py` (experiments); `a17_run.py`, `a17_driver.py`, `a17_go2.py` to `a17_go7.py` (job chains); `a17_lib.py`, `a17_stats.py`, `a17_figs.py` (analysis and figures); `lp_u17.py`, `a17_t1.py`, `a17_t2.py`, `a17_t2b.py`, `a17_t3.py` (this paper).